\documentclass{article}

\usepackage[utf8]{inputenc} 
\usepackage[T1]{fontenc}    
\usepackage{hyperref}       
\usepackage{url}            
\usepackage{booktabs}       
\usepackage{amsfonts}       
\usepackage{nicefrac}       
\usepackage{microtype}      

\newcommand{\Tau}{\mathrm{T}}
\usepackage{xcolor}         
\usepackage{makecell}
\usepackage{enumitem}
\usepackage{amsmath}
\usepackage{enumitem}
\usepackage{adjustbox}  
\usepackage{amssymb}

\usepackage{amsmath,amsfonts,bm}

\def\eqref#1{equation~\ref{#1}}

\def\1{\bm{1}}

\DeclareMathAlphabet{\mathsfit}{\encodingdefault}{\sfdefault}{m}{sl}
\SetMathAlphabet{\mathsfit}{bold}{\encodingdefault}{\sfdefault}{bx}{n}

\def\gG{{\mathcal{G}}}

\def\sA{{\mathbb{A}}}

\def\sK{{\mathbb{K}}}

\def\sV{{\mathbb{V}}}

\usepackage{caption}

\usepackage[ruled,linesnumbered]{algorithm2e}

\usepackage[square, numbers]{natbib}
\newif\ifanonymous
\anonymousfalse
\ifanonymous
    \usepackage[nonatbib]{neurips_2021}
\else
    \usepackage[preprint, nonatbib]{neurips_2021}
\fi

\title{ScheduleNet: Learn to solve multi-agent scheduling problems with reinforcement learning}

\author{
  Junyoung Park \\
  KAIST \\
  \texttt{junyoungpark@kaist.ac.kr} \\
  \And
  Sanjar Bakhtiyar \\
  KAIST \\
  \texttt{sanzhbakh@kaist.ac.kr} \\
  \And
  Jinkyoo Park \\
  KAIST \\
  \texttt{jinkyoo.park@kaist.ac.kr} \\
}

\begin{document}

\maketitle

\begin{abstract}
We propose ScheduleNet, a RL-based real-time scheduler, that can solve various types of multi-agent scheduling problems. We formulate these problems as a semi-MDP with episodic reward (makespan) and learn ScheduleNet, a decentralized decision-making policy that can effectively coordinate multiple agents to complete tasks. The decision making procedure of ScheduleNet includes: (1) representing the state of a scheduling problem with the agent-task graph, (2) extracting node embeddings for agent and tasks nodes, the important relational information among agents and tasks, by employing the type-aware graph attention (TGA), and (3) computing the assignment probability with the computed node embeddings. We validate the effectiveness of ScheduleNet as a general learning-based scheduler for solving various types of multi-agent scheduling tasks, including multiple salesman traveling problem (mTSP) and job shop scheduling problem (JSP).

\end{abstract}

\section{Introduction}
Optimal assignments of multiple autonomous agents for sequential completion of distributed tasks is necessary to solve various types of scheduling problems in logistics, transportation, and manufacturing. Finding the optimal delivery plans for vaccines, customer pickup order for ride sharing services, and machine operation sequence in modern manufacturing facilities are some examples of such scheduling problems. As the size of the problems increase, it is imperative to design an effective scheduler to solve these complex problems.

Solving large-scale scheduling problems using mathematical programming is infeasible/ineffective due to (1) the expensive computational cost, and (2) the inability to modify the scheduling action during real-time execution. As a remedy,  learning-based approaches, especially reinforcement Learning (RL), have been proposed to solve the traveling salesman problem (TSP) and the vehicle routing problem (VRP) \cite{bello2016neural, kool2018attention, khalil2017learning,nazari2018reinforcement}. Although recent works show the promising performances, they have been limited to solve only single-agent scheduling problems. RL-based approaches that solve multi-agent scheduling problems (mSP) are underrepresented in the research community, even though mSP poses greater scientific challenges and covers a broader set of real-world problems. 

\textbf{Objective}. In this paper, we propose ScheduleNet, a RL-based real-time decentralized multi-agent scheduler. ScheduleNet builds solution sequentially by accommodating for current partial solution and actions of other agents. ScheduleNet attempt to achieve following three goals:
\vspace{-0.3cm}
\begin{itemize}[leftmargin=*]
    \item \textit{General scheduler} that can effectively solve various types mSPs.
    \item \textit{Cooperative scheduler} that can effectively induce coordination among multiple agents to minimize the total completion time (makespan) of distributed tasks.
    \item \textit{Scalable scheduler} that can solve the large-scale mSPs in a computationally efficient way.
\end{itemize}

\textbf{Formulation}. To achieve these goals, we formulate mSPs as a semi Markov decision process (MDP) with episodic reward (e.g. makespan) and seek to derive a decentralized decision making policy that can be shared by all agents. In the proposed semi-MDP, the state is defined as the current status (partial solution) of a target mSP, and the action is defined as an assignment of an idle agent to one of the remaining and feasible tasks. The proposed formulation has following advantages:
\vspace{-0.25cm}
\begin{itemize}[leftmargin=*]
    \item \textit{Direct optimization of scheduling objective (makespan)}. This formulation alleviates the need for devising dense reward functions, which can be very challenging for complex problems and do not guarantee an optimal cooperative behavior of agents.
    \item \textit{Decentralization of scheduling policy that can be transferred to any sized problems}. ScheduleNet allows each agent chooses its destination independently while using its local observations and incorporating other agents' assignments. This decentralization ensures that the learned policy can solve problems with any number of agents and tasks in a scalable manner.
\end{itemize}

\textbf{Solution construction method}. ScheduleNet constructs the solution using a sequential decision-making framework. At every step, ScheduleNet accepts the MDP state as an input and assigns an idle agent to one of the feasible tasks. The decision-making procedure of ScheduleNet is as follows:
\vspace{-0.25cm}
\begin{itemize}[leftmargin=*]
    \item ScheduleNet first represents the MDP state as a agent-task graph, which is both effective in capturing complex relationships among the entities and general enough to be applied to various types of mSPs.
    \item ScheduleNet then employs the type-aware graph attention (TGA) to extract important relational features among agents and tasks in a computationally efficient manner.
    \item Lastly, ScheduleNet computes the agent-task assignment probability by utilizing the computed node embeddings.
\end{itemize}

\textbf{Training Method}. Although makespan (shared team reward) is the most direct and general reward design for solving mSPs, training a decentralized scheduling policy using this reward is extremely difficult due to the credit assignment issues \cite{riedmiller2018learning, hare2019dealing}. Additionally, makespan is highly volatile due to the combinatorial aspect of mSPs' solution space; a small change in a solution can drastically alter the outcome. Thus, training the decentralized scheduling policy is an extremely challenging task. To overcome these difficulties, we propose a RL training scheme, which empirically increases the stability of learning and asymptotic performance of ScheduleNet.

\textbf{Validation}. We validate the effectiveness of ScheduleNet on two types of mSPs: multiple traveling salesmen problem (mTSP) and jop-shop scheduling problem (JSP). From a series of experiments on random mSP instances and benchmark datasets, we empirically show that ScheduleNet can outperform other heuristic approaches and existing deep RL approaches. Furthermore, ScheduleNet demonstrates its exceptional effectiveness on large and practical problems.

\section{Related Works}
\textbf{Solving single-agent routing (scheduling) problems with RL}. According to \citep{mazyavkina2020reinforcement}, the RL approaches to solving agent routing problems can be categorized into: (1) improvement heuristics learns to rewrite the complete solution iteratively to obtain a better solution \citep{wu2020learning, pmlr-v129-costa20a, NEURIPS2019_131f383b, Lu2020A}; (2) construction approach learns to construct a solution by sequentially assigning idle agents to unvisited cities until the full routing schedule (sequence) is constructed \citep{bello2016neural, nazari2018reinforcement, kool2018attention, khalil2017learning}, and (3) hybrid approaches blending both approaches \citep{joshi2020learning, fu2021generalize, kool2021deep, pmlr-v119-ahn20a}. Typically, learning-based improvement or hybrid approaches have shown good performance since these can iteratively update the best solution until reaching the best one. However, these approaches usually require a longer computational time. In addition, since these approaches run in a centralized manner, they are hard to be expanded to multi-agent seating where agents need to be operated in a decentralized manner.

\textbf{Learned mTSP solvers}. There are only few RL approaches for solving mTSP. \cite{kaempfer2018learning} proposed a variant of transformer \cite{vaswani2017attention} that trains the policy assigning a agent to a city using the mTSP solutions computed by integer linear programming solvers (imitation learning). \cite{hu2020reinforcement} applies RL to train the clustering algorithm to group cities and applied (1) trained RL-policy for solving TSP or (2) strong TSP heuristics (OR-Tool) to optimize the sub-tour in each cluster. Being different from these approaches, we focus on deriving a complete \textit{end-to-end} learned heuristic that constructs a feasible solution from ``scratch'' without relying on any existing solvers. For this reason, We did not consider \cite{kaempfer2018learning}, and \cite{hu2020reinforcement} as baseline algorithms since they use 2-stage approaches, relying on learned or well known TSP heuristics.

\textbf{Learned JSP solvers}. Some RL methods have also been proposed for solving JSP. For example, \cite{gabel2012distributed, lin2019smart} have proposed to learn scheduling policy for each agent and hence, it requires an additional training to solve JSPs with a different number of agents from the training cases. Recently, \cite{Park2021learning, zhang2020learning} have proposed to learn a shared scheduling policy for all agents while utilizing the disjunctive graph representation of JSP.
Unlike these methods utilizing well-designed dense reward, we directly use the makespan reward to train a policy.

\section{Problem Formulation}
We formulate mSP as a semi-MDP with sparse reward and aim to derive a decentralized scheduling decision-making policy that can be shared by all agents. The semi-MDP is defined as:

\textbf{State}. We define state $s_\tau$ as the $\tau$-th partial solution of mSP (i.e., the completed/uncompleted tasks, the status of agents, and the sequence of the past assignments). The initial $s_0$ and terminal state $s_\Tau$ are defined as an empty and a complete solution respectively.

\textbf{Action}. We define action $a_\tau$ as the act of assigning an idle agent to one of the feasible tasks (unassigned tasks). We refer to $a_\tau$ as the \textit{agent-to-task assignment}.\footnote{When the multiple agents are idle at the same time $t$, we randomly choose one agent and assign an action to the agent and repeat the process until no agent is idle. Note that such randomness do not alter the resulting solutions, since the agents are considered to be homogeneous and the scheduling policy is shared.}


\textbf{Transition}. The proposed MDP is formulated with an \textit{event}-based transition. An event is defined as the the case where any agent finishes the assigned task (e.g. a salesman reaches the assigned city in mTSP). Whenever an event occurs, the idle agent is assigned to a new task, and the status of the agent and the target task are updated accordingly. We enumerate the event with $\tau$ to avoid confusion from the elapsed time of the problem; $t(\tau)$ is a function that returns the time of event $\tau$.

\textbf{Reward}. The proposed MDP uses the minus of makespan (i.e. total completion time of tasks) as a reward, i.e., $r({s_\Tau}) = -t(\Tau)$, that is realized only at $s_\Tau$.

\begin{figure*}[t]
\begin{center}
\includegraphics[width={1.0\linewidth}]{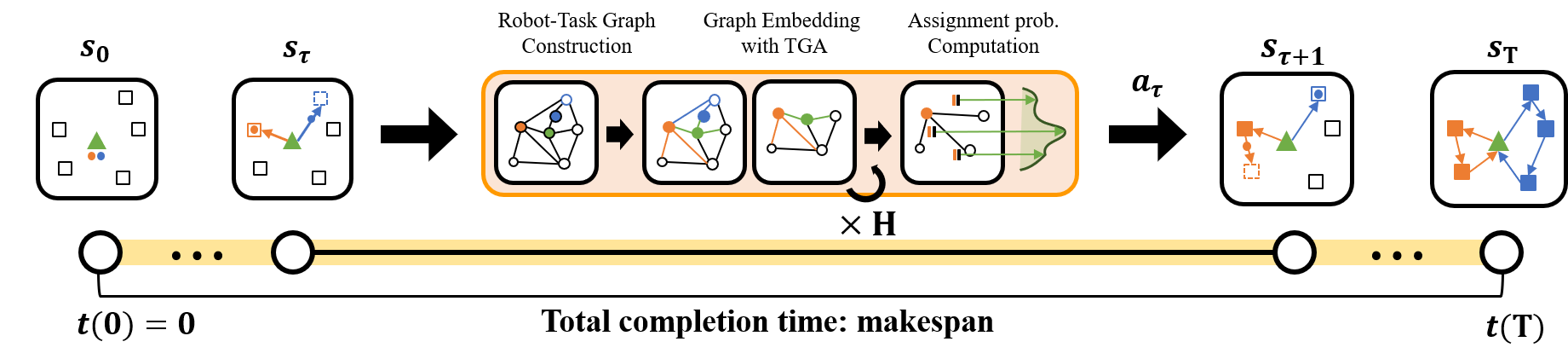}
\end{center}
\caption{\textbf{Solving mSP with ScheduleNet.} At every event of the semi-MDP, ScheduleNet constructs the agent-task graph $\gG_\tau$ from $s_\tau$, then computes the node embedding of $\gG_\tau$ using TGA, and finally computes the agent-task assignment probabilities from the node embedding.}
\label{fig:schedulenet}
\vspace{-0.2cm}
\end{figure*}

\subsection{Example: MDP formulation of mTSP}
\label{sec:mtps-mdp-formulation}
Let us consider the single-depot mTSP with two types of entities: $m$ salesmen (i.e. $m$ agents) and $N$ cities (i.e. $N$ tasks) to visit. All salesmen start their journey from the depot and come back to the depot after visiting all cities (each city can be visited by only one salesman). A solution to mTSP is considered to be \textit{complete} when all the cities have been visited and all salesmen  have returned to the depot. The semi-MDP formulation for mTSP is similar to that of the general mSP. The specific definitions of the state and reward for mTSP are as follows:

\textbf{State.} We define $s_{\tau}=(\{s_{\tau}^{i}\}_{i=1}^{N+m}, s_{\tau}^{\text{env}})$ is composed of two types of states: entity state $s_{\tau}^{i}$ and environment state $s_{\tau}^{\text{env}}$.
\vspace{-0.35cm}
\begin{itemize}[leftmargin=*]
\setlength\itemsep{-0.1em}
    \item $s_{\tau}^{i} = (p_{\tau}^{i}, \1_{\tau}^{\text{active}},\1_{\tau}^{\text{assigned}})$ is the state of $i$-th entity. $p_{\tau}^{i}$ is the position of $i$-th entity at the $\tau$-th event. $\1_{\tau}^{\text{active}}$ indicates whether the $i$-th worker/task is active (worker is working/ task is not visited) or not. Similarly, $\1_{\tau}^{\text{assigned}}$ indicates whether worker/task is assigned or not.
    \item $s_{\tau}^{\text{env}}$ contains the current time of the environment, and the sequence of cities visited by each salesman.
\end{itemize}
\vspace{-0.35cm}

\textbf{Reward.} We use makespan of mTSP as a reward, which is sparse and episodic. The reward $r({s_\tau}) = 0$ for all non-terminal events, and $r({s_\Tau}) = -t(\Tau)$, where $\Tau$ is the index of the terminal state.

\section{ScheduleNet}
In this section, we explain how ScheduleNet recommends a  scheduling action $a_\tau$ of an idle agent from input $s_\tau$ (partial solution) by (1) constructing the agent-task graph $\gG_\tau$, (2) embedding $\gG_\tau$ using TGA, and (3) computing the assignment probabilities. Figure \ref{fig:schedulenet} illustrates the decision-making process of ScheduleNet.

\subsection{Constructing agent-task graph}
\label{sec:graph-representation} 
ScheduleNet constructs the \textit{agent-task graph} $\gG_\tau$ that reflects the complex relationships among the entities in $s_\tau$.
Specifically, ScheduleNet constructs a directed complete graph $\gG_\tau = (\sV, \mathbb{E})$ out of $s_{\tau}$, where $\sV$ is the set of nodes and $\mathbb{E}$ is the set of edges. The nodes and edges, and their associated features are defined as:
\begin{itemize}[leftmargin=*]
\vspace{-0.35cm}
\setlength\itemsep{-0.1em}
    \item $v_i$ denotes the $i$-th node representing either a agent or a task. $v_i$ contains the node feature $x_i=(s_{\tau}^i,k_i)$, where $s_{\tau}^i$ is the state of entity $i$, and $k_i$ is the type of $v_i$. For example, if the entity $i$ is \textit{agent} and its $\1_\tau^{active}=1$, then the $k_i$ becomes \textit{active-agent} type. For the full list of the node types, refer to Appendix \ref{appendix:mtsp-graph-formulation}.
    \item $e_{ij}$ denotes the edge between the source node $v_j$ and the destination node $v_i$. The edge feature $w_{ij}$ is equal to the Euclidean distance between the two nodes.
\end{itemize}
In the following subsections, we omit the event iterator $\tau$ for notational brevity, since the action selection procedure is only associated with the current event index $\tau$. 

\subsection{Graph embedding using TGA}
ScheduleNet computes the node embeddings from the agent-task graph $\gG$ using the type-aware graph attention (TGA). The embedding procedure first encodes the features of $\gG$ into the initial node embeddings $\{h_i^{(0)}|v_i \in \sV \}$, and the initial edge embeddings $\{h_{ij}^{(0)}|e_{ij} \in \mathbb{E}\}$. ScheduleNet then performs TGA embedding $H$ times to produce final node embeddings $\{h_i^{(H)}|v_i \in \sV \}$, and edge embeddings $\{h_{ij}^{(H)}|e_{ij} \in \mathbb{E}\}$. To be specific, TGA embeds the input graph using the type-aware edge update, type-aware message aggregation, and the type-aware node update as explained in the following paragraphs.

\textbf{Type-aware edge update}. Given the node embedding ${h_i}$ and edge embedding $h_{ij}$, TGA computes the type-aware edge embedding $h'_{ij}$ and the attention logit $z_{ij}$ as follows:
\begin{align}
\label{eqn:gnn-edge-context}
\begin{split}
h'_{ij} &= \text{TGA}_{\mathbb{E}} ([h_i, h_j, h_{ij}], k_j) \\
z_{ij} &= \text{TGA}_{\mathbb{A}} ([h_i, h_j, h_{ij}], k_j) \\
\end{split}
\end{align}
where $\text{TGA}_{\mathbb{E}}$ and $\text{TGA}_{\mathbb{A}}$ are the type-aware edge update function and the type-aware attention function, respectively.

$\text{TGA}_{\mathbb{E}}$ and $\text{TGA}_{\mathbb{A}}$ are parameterized as Multilayer Perceptron (MLP) where the first layer is the Multiplicative Interaction (MI) layer \citep{jayakumar2019multiplicative}. The MI layer, which is a bilinear instantiation of hypernetwork \citep{ha2016hypernetworks}, adaptively generates parameters of $\text{TGA}_{\mathbb{E}}$ and $\text{TGA}_{\mathbb{A}}$  based on the type $k_j$. This allows us to use $\text{TGA}_{\mathbb{E}}$ and $\text{TGA}_{\mathbb{A}}$ for all types of nodes, and thus reduces the number of embedding functions to be learned while maintaining the good representational power of GNN. 

\textbf{Type-aware message aggregation}. Each entity in the agent-task graph interacts differently with the other entities, depending on the type of the edge between them. To preserve the different relationships between the entities during the graph embedding procedure, TGA gathers messages $h_{ij}'$ via the type-aware message aggregation.


First, TGA aggregates messages for each node type (\textit{per-type}) and produces the \textit{per-type} message $m_{i}^k$ as follows:
\begin{align}
\label{eqn:tga-intra-aggr}
\begin{split}
m_{i}^k = \sum_{j \in \mathcal{N}_k(i)}\alpha_{ij} h'_{ij}
\end{split}
\end{align}
where $\mathcal{N}_k(i) = \{v_l| k_{l} = k, \forall v_l \in \mathcal{N}(i)\}$ is the type $k$ neighborhood of $v_i$, and $\alpha_{ij}$ is the attention score that is computed using $z_{ij}$:
\begin{align}
\label{eqn:tga-attention}
\begin{split}
\alpha_{ij} &= \frac{\text{exp}(z_{ij})}{\sum_{j \in \mathcal{N}_k(i)}\text{exp}(z_{ij})}  \\
\end{split}
\end{align}

TGA then concatenates the per-type messages to produce the aggregated message $m_i$ as:
\begin{align}
\label{eqn:tga-inter-aggr}
\begin{split}
m_i &= \text{concat}(\{m_{i}^k | k \in  \sK\}) \\
\end{split}
\end{align}
where $\sK$ is the set of node types. Since the number of node types is fixed, the size of $m_i$ is fixed regardless of the size of problems.

\textbf{Type-aware node update}. 
The aggregated message $m_i$ is then used to compute the updated node embedding $h_i'$ as follows:
\vspace{-0.35cm}
\begin{align}
\label{eqn:gnn-edge-context}
h'_{i} &= \text{TGA}_{\mathbb{V}} ([h_i, m_i], k_i)
\end{align}

where $\text{TGA}_{\mathbb{V}}$ is the type-aware node update function that is parametrized with MLP. The first layer of $\text{TGA}_{\mathbb{V}}$ is MI Layer, similar to the edge updater. The detailed architectures of $\text{TGA}_{\mathbb{E}}$, $\text{TGA}_{\mathbb{A}}$, and $\text{TGA}_{\mathbb{V}}$ are provided in Appendix \ref{appendix:TGA}.

\subsection{Computing assignment probability}
Using the computed final node embeddings $\{h_i^{(H)}|v_i \in \sV \}$ and edge embeddings $\{h_{ij}^{(H)}|e_{ij} \in \mathbb{E}\}$, ScheduleNet selects the best assignment action $a_\tau$ for the target agent.

ScheduleNet computes the assignment probability of the target \textit{idle} agent $i$ to the \textit{unassigned} task $j$ as follows:
\begin{align}
\label{eqn:policy}
\begin{split}
l_{ij} &= \text{MLP}_{\textit{actor}}(h^{(H)}_i, h^{(H)}_j, h^{(H)}_{ij}) \\
p_{ij} &= \text{softmax}(\{l_{ij}\}_{j \in \sA(\gG_\tau)})
\end{split}
\end{align}
where  $h^{(H)}_i$ is the final node embedding for $v_i$, and $h^{(H)}_{ij}$ is the final edge embeddings for $e_{ij}$. In addition, $\sA(\gG_\tau)$ denote the set of feasible actions defined as $\{v_j| k_j=\textit{Unassigned-task}\, \forall j \in \sV \}$.

Note that ScheduleNet learns how to process local information of the state and make the decentralized action for each agent. This allows ScheduleNet to compute the assignment probabilities for mSPs with \textit{arbitrarily} numbered agents and tasks in a \textit{scalable} manner.

\section{Training ScheduleNet}
We utilize sparse team reward (makespan) as the direct reward signal to train the decentralized scheduler (ScheduleNet) for having multiple agents to complete tasks as quickly as possible. Even though this team reward is the most direct signal that can be used for solving various types of mSPs, training a decentralized cooperative policy using a single sparse and delayed reward is notoriously difficult \cite{riedmiller2018learning, hare2019dealing}. The high variance of the reward signal due to the combinatorial nature of mSPs' solution space adds an additional difficulty. To handle such difficulties, we employ two training stabilizers, reward normalization, and Clip-REINFORCE, while training ScheduleNet.

\subsection{Reward normalization}
We denote the makespan induced by policy $\pi_{\theta}$ as $M(\pi_{\theta})$. We observe that $M(\pi_{\theta})$ is highly volatile depending on the problem size ($N$, $m$) and $\pi_{\theta}$. To reduce the variance of the reward incurred from the problem sizes, we propose to use the normalized makespan $\bar{M}(\pi_{\theta}, \pi_b)$ computed as:
\begin{align}
\label{eqn:inter-problem-norm}
\begin{split}
\bar{M}(\pi_{\theta}, \pi_{b}) =  \frac{M(\pi_{\theta})-M(\pi_b)}{M(\pi_{b})}\
\end{split}
\end{align}
where $\pi_b$, a baseline policy, is the current policy $\pi$ in greedy (test) mode. $\bar{M}(\pi_{\theta}, \pi_b)$ measures the relative scheduling performance of $\pi_{\theta}$ to $\pi_b$. Obviously,  $\bar{M}(\pi_{\theta}, \pi_b)$ has smaller variance than $M(\pi_{\theta})$, especially when the number of agent $m$ changes.

Using $\bar{M}(\pi_{\theta}, \pi_b)$, we compute the normalized return $G_\tau(\pi_{\theta}, \pi_b)$ as follows:
\begin{align}
\label{eqn:return}
\begin{split}
G_\tau(\pi_{\theta}, \pi_b) \triangleq -\gamma^{T-\tau} \bar{M}(\pi_{\theta}, \pi_b)
\end{split}
\end{align}
where $T$ is the index of the terminal state, and $\gamma$ is the discount factor of MDP. The minus sign is for minimizing the makespan. Note that, in the early phase of mSP (when $\tau$ is small), it is difficult to estimate the makespan. Thus, we place a smaller weight (i.e, $\gamma^{T-\tau}$) for $G_\tau$ evaluated when $\tau$ is small.

\subsection{Clip-REINFORCE}

Even a small change in a single assignment can result in a dramatic change to make span due to the combinatorial nature of the solution. Hence, training value function that predicts the $G_\tau$ reliably is difficult. See Appendix \ref{appendix:ablation-studies} for more information. We thus propose to utilize the Clip-REINFORCE (CR), a variant of PPO \cite{schulman2017proximal}, \textit{without} the learned value function for training ScheduleNet.  The objective of the Clip-REINFORCE is given as follows: 
\begin{flalign}
    &\mathcal{L}(\theta) = \mathop{\mathbb{E}}_{(\gG_\tau, a_\tau) \sim \pi_\theta}[\text{min}(\text{clip}(\rho_\tau, 1-\epsilon, 1+\epsilon) G_{\tau}, \rho_\tau G_{\tau})]
\end{flalign}
where $G_{\tau}$ is a shorthand notation for $G_{\tau}(\pi_\theta, \pi_b)$ and $\rho_\tau = \pi_\theta(a_\tau|\gG_\tau)/\pi_{b}(a_\tau | \gG_\tau)$ is the ratio between the target and baseline policy. 

\begin{figure}[t]
   \begin{minipage}{0.7\textwidth}
    \centering
    \includegraphics[width=.99\linewidth]{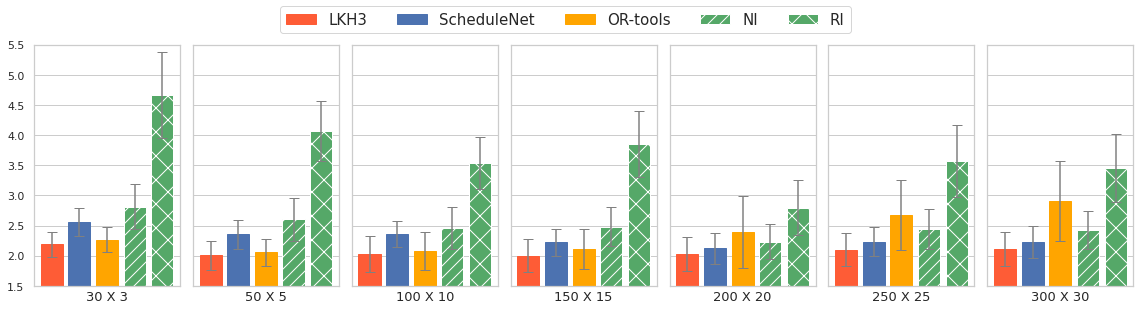}
    \caption{\textbf{Scheduling performance on random ($N \times m$) mTSP datasets (smaller is better)}. The y-axis shows the normalized makespans. The red, blue and orange bar charts demonstrate the performance of LKH3, SchdeduleNet and OR-tools respectively. The green bars show the performance of two-phase heuristics. The error bars shows $\pm 1.0$ standard deviation of the makespans.}
    \label{fig:makesapn_on_random}
    \end{minipage}\hfill
    \begin{minipage}{0.25\textwidth}
    \centering
    \includegraphics[width=.99\linewidth]{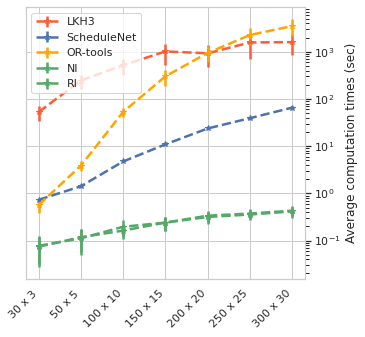}
    \caption{\textbf{Inference time of ScheduleNet and the baselines}}
    \label{fig:comp_time_on_random}
   \end{minipage}
\end{figure}

\section{Experiments}
In this section, we evaluate the performance of ScheduleNet on mTSP and JSP. To calculate the inference time, we run all experiements on the server equipped with a NVIDIA Titan X GPU and AMD Threadripper 2990WX CPU. we use the GPU to evaluate ScheduleNet, and single CPU core for all other baseline algorithms. In the testing phase, the action with the highest probability is selected greedily and used. No further solution refinement techniques such as beam search and random shooting, are applied to test ScheduleNet.

\textbf{Performance metrics}. It is computationally infeasible to obtain the optimal solution for mTSP with MILP solvers (e.g. CPLEX) even for the small-sized problems. 
Therefore, when the optimal solution is unknown, we define the \textit{scheduling performance} of scheduler $\pi$ as $M(\pi)$. On the other hand, when the optimal solution is known, we report the scheduling performance as the optimality gap, which is defined as $M(\pi)/M^*$, where $M^*$ is the optimal makespan.

\subsection{mTSP experiments}
\label{section:mtsp-experiments}
\textbf{Training.} We denote ($N \times m$) as the mTSP with $N$ cities (tasks) and $m$ salesmen (agents). To generate a random mTSP instance, we sample $N$ and $m$ from $U(20, 25)$ and $U(2, 5)$, respectively. Similarly, the Euclidean coordinates of $N$ cities are sampled uniformly at random from the unit square. ScheduleNet is trained on random mTSP instances that are generated on-the-fly. For all experimental results, we evaluate the performance of a single trained ScheduleNet model. Please refer to Appendix \ref{appendix:mtsp-training} for more information.

\textbf{Results on random mTSP datasets}. To investigate the generalization capacity of ScheduleNet to various problem sizes, we evaluate ScheduleNet on the randomly generated mTSP datasets. Each ($N \times m$) dataset consists of 100 random uniform mTSP instances. 

The baseline algorithms we considered are: 
\vspace{-0.2cm}
\begin{itemize}[leftmargin=*]
\setlength\itemsep{-0.1em}
    \item Lin-Kernighan-Helsgaun 3 (LKH3) \cite{helsgaun2017extension}, a state-of-the-art heuristic solver, that are known to find the optimal solutions for the mTSP problems with the identified optimal solutions. We use the makespans computed by LKH3 as the proxies for the optimal solutions of the test mTSP instances.
    \item Google OR-Tools\cite{ortools}, a highly optimized meta-heuristic library.
    \item NI and RI, two-phase mTSP heuristics, that first group the cities with $K$-means clustering (where $K=m$) and then apply the well-known TSP heuristics, i.e., Nearest Insertion (NI) and Random Insertion (RI), per cluster.
\end{itemize}


From Figure \ref{fig:makesapn_on_random}, in the small-sized maps ($N < 200$), we can notice that the scheduling performances of ScheduleNet better than that of the two-phase heuristics and worse than that of OR-tools. In the larger maps ($N \geq 200$), on the other hand, ScheduleNet outperforms OR-tools and the two-phase heuristics as well. ScheduleNet exhibits small standard deviations of makespans for all problem sizes. In addition, Figure \ref{fig:comp_time_on_random} compares the average inference time and the standard deviation per problem size, clearly indicating that ScheduleNet is significantly faster especially for large size mTSP instances.


\textbf{Results on Public Benchmarks}. Next, to explore the generalization of ScheduleNet to problems that come from completely different distributions (i.e. real-world data), we present the results on the mTSPLib dataset defined by \citet{necula2015tackling}. mTSPLib consists of four small-to-medium- sized problems from TSPLib \citep{reinelt1991tsplib}, each of which extended to multi-agent setup where $m$ equals to 2, 3, 5, and 7.

The baseline algorithms are: (1) exact solver CPLEX, (2) LKH3 whose stopping criteria set to be the known best solutions, (3) OR-Tools, (4) population-based meta-heuristics Self-organization map (SOM), Ant-colony optimization (ACO), Evolutionary algorithm (EA) \citep{lupoaie2019som}, and (5) the two-phase heuristics (NI and RI). From Table \ref{table:mtsp-benchmark-table-transpose}, we can observe that, at small-to-medium scale, OR-Tools produces near-optimal solutions, followed by ScheduleNet and ACO. However, the two-phase heuristics show drastic performance degradation, which is most likely due to the non-uniform distribution of the cities. From this experiment, we can observe that the ScheduleNet policy is equally effective in solving both generated and real-world mTSPs problems.

\begin{table*}[t]
\centering
\renewcommand{\arraystretch}{1.3} 
\caption{\textbf{mTSPLib results.} The CPLEX results with $*$ are optimial solutions. Otherwise, the known-best upper bound of CPLEX results are reported \cite{mTSPLib}. Population based metaheuristics results are reproduced from \citet{lupoaie2019som}.}
\vskip 0.15in
\resizebox{\textwidth}{!}{
\begin{tabular}{c|cccc|cccc|cccc|cccc|c}
\toprule
Instance    & \multicolumn{4}{c|}{\textit{eil51}}    & \multicolumn{4}{c|}{\textit{berlin52}}     & \multicolumn{4}{c|}{\textit{eil76}}    & \multicolumn{4}{c|}{\textit{rat99}}    & \multicolumn{1}{c}{Gap} \\
\midrule
$m$         & 2     & 3     & 5     & 7     & 2      & 3      & 5      & 7      & 2     & 3     & 5     & 7     & 2     & 3     & 5     & 7     &                          \\
\midrule
CPLEX       & 222.7$^*$ & 159.6 & 124.0 & 112.1 & 4110.2 & 3244.4 & 2441.4 & 2440.9 & 280.9$^*$ & 197.3 & 150.3 & 139.6 & 728.8 & 587.2 & 469.3 & 443.9 & 1.00 \\ \hline
LKH3       & 222.7 & 159.6 & 124.0 & 112.1 & 4110.2 & 3244.4 & 2441.4 & 2440.9 & 280.9$^*$ & 197.3 & 150.3 & 139.6 & 728.8 & 587.2 & 469.3 & 443.9 & 1.00 \\ \hline
OR-Tools    & 243.3 & 170.5 & 127.5 & 112.1 & 4665.5 & 3311.3 & 2482.6 & 2440.9 & 318.0 & 212.4 & 143.4 & 128.3 & 762.2 & 552.1 & 473.7 & 442.5 & 1.03                     \\ \hline
ScheduleNet & 259.7 & 172.2 & 118.9 & 112.4 & 4816.3 & 3372.1 & 2615.6 & 2576.0 & 334.1 & 226.5 & 168.0 & 151.3 & 790.0 & 579.3 & 502.5 & 471.7 & 1.08                     \\
\midrule
SOM         & 278.4 & 210.3 & 157.7 & 136.8 & 5350.8 & 4197.6 & 3461.9 & 3125.2 & 364.0 & 278.6 & 210.7 & 183.1 & 927.4 & 756.1 & 624.4 & 564.1 & 1.31                     \\ 
ACO         & 248.8 & 180.6 & 135.1 & 120.0 & 4389.0 & 3468.9 & 2733.6 & 2510.1 & 308.5 & 224.6 & 163.9 & 146.9 & 767.2 & 620.5 & 525.5 & 492.1 & 1.09                     \\ 
EA          & 276.6 & 208.2 & 151.2 & 123.9 & 5038.3 & 3865.5 & 2853.6 & 2543.7 & 365.7 & 285.4 & 211.9 & 177.8 & 896.7 & 739.4 & 596.9 & 534.9 & 1.24                     \\ 
\midrule
NI          & 271.3 & 202.9 & 183.5 & 129.7 & 5941.0 & 3811.5 & 2972.6 & 2972.6 & 363.2 & 302.1 & 191.4 & 173.8 & 916.6 & 802.8 & 668.6 & 554.2 & 1.30                     \\
RI          & 265.9 & 195.9 & 150.5 & 127.7 & 5785.0 & 4133.9 & 4108.6 & 2998.2 & 395.2 & 276.6 & 185.8 & 155.5 & 890.9 & 843.0 & 675.4 & 565.1 & 1.31                     \\
\bottomrule
\end{tabular}
}
\label{table:mtsp-benchmark-table-transpose}
\vspace{-0.2cm}
\end{table*}

\begin{figure}[!htb]
    \centering
    \begin{minipage}{.475\textwidth}
        \flushleft
        \includegraphics[width={1\linewidth}]{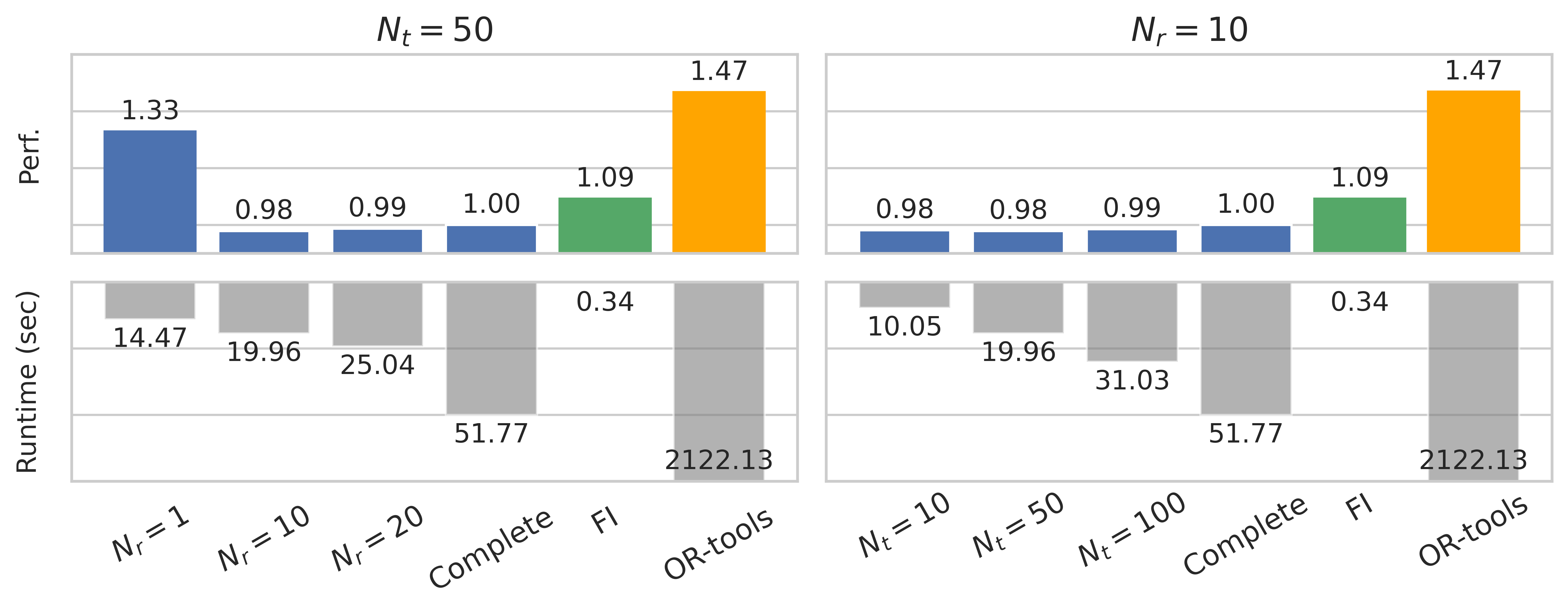}
        \caption{\textbf{Scheduling performance (top) and computational time (bottom) on the sparse graphs.} The left figure shows the scheduling performance w.r.t. $N_r$. The right figure w.r.t. $N_t$. We normalize the makespan of each scheduling algorithm with the ScheduleNet makespan with complete graph.}
        \label{fig:knn-graph-ablation}
    \end{minipage}%
    \hspace{0.5cm}
    \begin{minipage}{0.475\textwidth}
        \flushright
        \includegraphics[width={1.0\linewidth}]{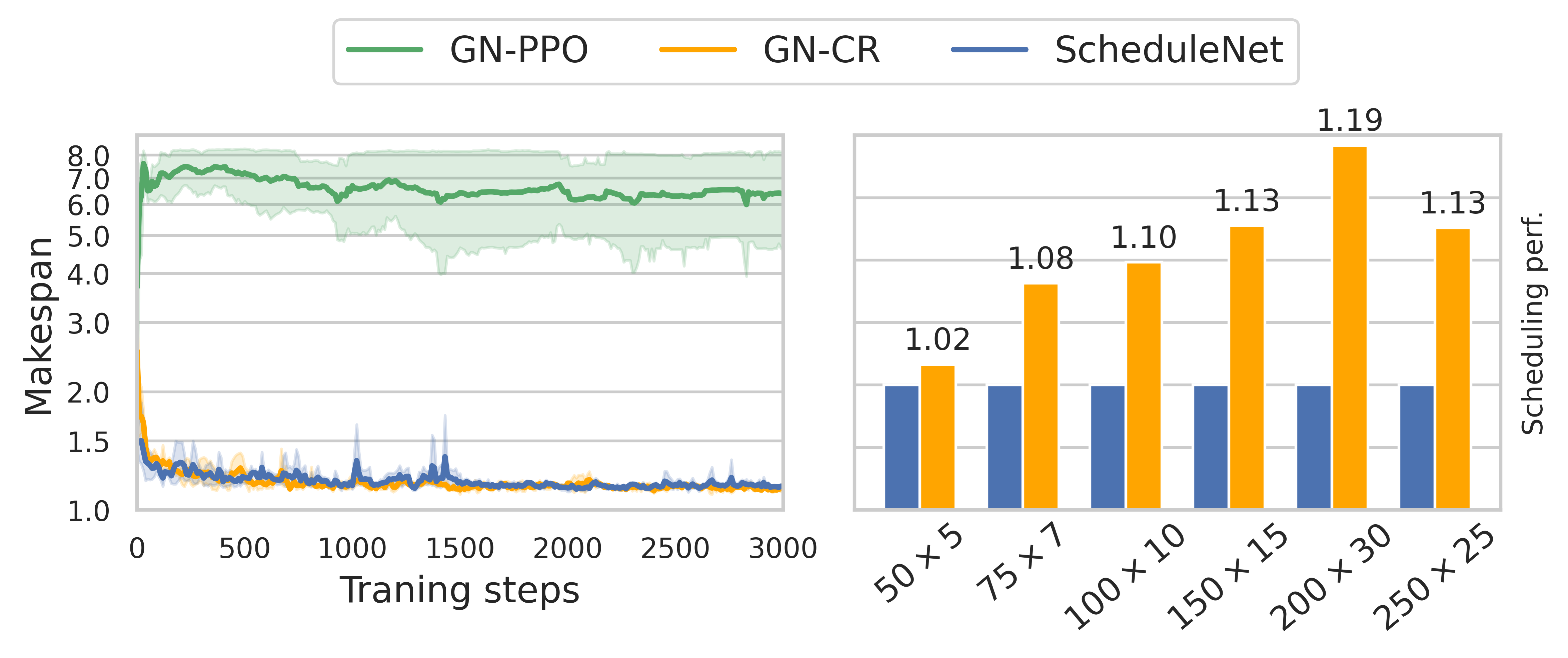}
        \caption{\textbf{Training ablation.} Makespan over the training steps (left) and scheduling performance on uniform random test datasets (right). The blue, orange, and green curves show the normalized makespan of {\fontfamily{lmtt}\selectfont ScheduleNet}, {\fontfamily{lmtt}\selectfont CR-GN}, and {\fontfamily{lmtt}\selectfont PPO-GN} resepctively.}
        \label{fig:training-ablation}
    \end{minipage}
\vspace{-0.2cm}
\end{figure}

\textbf{Graph sparsity ablation}. 
From the aforementioned performance results, we can conclude that ScheduleNet is specifically effective in solving large scale problems. However, as the problem size increases, the computational complexity for processing a complete graph will also increase in $\mathcal{O}(N^2+m^2)$. We hypothesize that a smaller value of $N_r$ (number of agent connection) and $N_t$ (number of task connection) reduce the computational cost significantly without losing performance a lot. To verify this idea, we evaluate the same ScheduleNet that is trained on small complete graphs (random mTSPs) on large mTSP instances ($300 \times 30$). 
The graph embedding procedure for the described sparse graph has complexity $\mathcal{O}((m+N)(N_r + N_t))$.

Figure \ref{fig:knn-graph-ablation} shows the scheduling performance (colored) and the computational time (gray) of ScheduleNet for different values of $N_r$ and $N_t$, i.e. the level of sparsity. The results show that the performance of the scheduleNet does not deteriorate significantly even with the limited information scope. These results imply that ScheduleNet can be used to solve very large real-world mTSPs in a fast and computationally efficient manner with local information. We also evaluate ScheduleNet on random mTSPs of size ($500 \times 50$) and ($750 \times 70$) (see Appendix \ref{appendix:mtsp-random-results}).

\textbf{RL training ablation}. To investigate the effectiveness of each ScheduleNet component (TGA and the training stabilizers), we train the following ScheduleNet variants:
\vspace{-0.2cm}
\begin{itemize}[leftmargin=*]
    \item {\fontfamily{lmtt}\selectfont GN-PPO} is a variant of ScheduleNet that has Attention-GN layers\footnote{AttentionGN does not use MI layer uses different message aggregation scheme. AttentionGN computes attention scores w.r.t. \textit{all incoming edges}, in type agnostic manner. In contrast, TGA attention scores are computed w.r.t. edge types with inter-type and intra-type aggregations. For more details about AttentionGN, refer to Algorithm 3 in Appendix D.} for graph embedding (instead of TGA) and is trained by the PPO.
    \item {\fontfamily{lmtt}\selectfont GN-CR} is a variant of ScheduleNet that has Attention-GN layers and is trained by Clip-REINFORCE with the normalized makespan.
    \item {\fontfamily{lmtt}\selectfont TGA-CR} is the ScheduleNet setting.
\end{itemize}

Figure \ref{fig:training-ablation} (left) visualizes the validation performance of ScheduleNet and its variants over the training steps.
It can be seen that {\fontfamily{lmtt}\selectfont GN-PPO} fails to learn meaningful scheduling policy, which indicates that solving mSP with the standard Actor-Critic methods can be a challenging. We see the same trend for TGA-PPO (result is omitted). In contrast, the model trained with the stabilizers ({\fontfamily{lmtt}\selectfont GN-CR} and {\fontfamily{lmtt}\selectfont ScheduleNet}) exhibit stable training, and converge to a similar performance level on the validation dataset.
In addition, as shown in Figure \ref{fig:training-ablation} (right), TGA-CR ({\fontfamily{lmtt}\selectfont ScheduleNet}) consistently shows better performance than {\fontfamily{lmtt}\selectfont GN-CR} over the random test datasets, showing that TGA plays an essential role in generalizing to larger mTSPs.



\subsection{JSP experiments}
We employ ScheduleNet to solve JSP, another type of mSP, to evaluate its generalization capacity in terms of solving various types of mSPs. The goal of the JSP is to schedule machines (i.e. agents) in a manufacturing facility to complete the jobs that consist of a series of operations (tasks), while minimizing the total completion time (makespan). Solving JSP is considered to be challenging since it imposes additional constraints that scheduler must obey: precedence constraints (i.e. an operation of a job cannot be processed until its precedence operation is done) and agent-sharing constraints (i.e. each agent has a unique set of feasible tasks). 

\textbf{Formulation}. We formulate JSP as a semi-MDP where $s_\tau$ is the partial solution of JSP, $a_\tau$ is agent-task assignment (i.e. assigning an idle machine to one of the feasible operations), and reward is the minus of makespan. Please refer to Appendix \ref{appendix:jsp-mdp-formulation} for the detailed formulation of semi-MDP and agent-task graph construction.

\textbf{Training}. We train ScheduleNet using the random JSP instances ($N \times m$) that have $N$ jobs and $m$ machines.
We sample $N$ and $m$ from $U(7, 14)$ and $U(2, 5)$ respectively to generate the training JSP instances. We randomly shuffle the order of the machines in a job to generate machine sharing constraints. Please refer to Appendix \ref{appendix:jsp-training} for more information.

\begin{figure*}[t]
\begin{center}
\includegraphics[width={1.0\linewidth}]{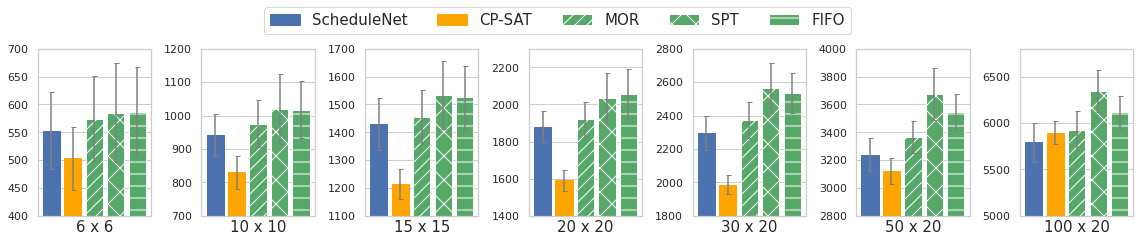}
\end{center}
\caption{\textbf{Scheduling Performance on random ($N \times m$) JSP datasets (smaller is better).} Each dataset contains 100 instances. The blue and orange bar charts demonstrate the performance of SchdeduleNet and CP-SAT respectively. The green bars show the performance of JSP heuristics.}
\label{fig:jsp-random}
\vspace{-0.2cm}
\end{figure*}

\textbf{Results on random JSP datasets}. To investigate the generalization capacity of ScheduleNet to various problem sizes, we evaluate ScheduleNet on the randomly generated JSP datasets. We compare the scheduling performance with the following baselines:  exact solver CP-SAT \cite{ortools} with one hour time-limit, and three JSP heuristics that known to work well in practice, Most Operation Remaining (MOR), First-in-first-out (FIFO), and Shortest Processing Time (SPT). As shown in Figure \ref{fig:jsp-random}, ScheduleNet always outperforms the baseline heuristics for all sizes of the maps and even outperforms CP-SAT for the larger maps. The result confirms that ScheduleNet has good generalization capacity in terms of the problem size.

\textbf{Results on Public Benchmarks}. We evaluate the scheduling performance to verify ScheduleNet's generalization capacity to unseen JSP distributions on the Taillard's 80 dataset \citep{taillard1993benchmarks}. We compare against two deep RL baselines \cite{Park2021learning, zhang2020learning}, as well as the mentioned heuristics (MOR, FIFO, and SPT). Both baseline RL methods were specifically designed to solve JSP by utilizing the well-known disjunctive JSP graph representation \cite{roy1964problemes} and well-engineered dense reward functions. Nevertheless, we can observe that ScheduleNet outperforms all baselines in all of the cases while utilizing only a sparse episodic reward (see Table \ref{table:jsp-benchmark-table}). Please refer to Appendix \ref{appendix:jsp-benchmark-results} for the extended benchmark results, including ORB \cite{applegate1991computational}, SWV \cite{storer1992new}, FT \cite{fisher1963probabilistic}, LA \cite{lawrence1984resouce}, YN \cite{yamada1992genetic}.

\begin{table}[t]
\centering
\scriptsize\addtolength{\tabcolsep}{-1pt}
\renewcommand{\arraystretch}{1} 
\caption{\textbf{The average scheduling gaps of ScheduleNet and the baseline algorithms.} The gaps are measured with respect to the optimal (or the best-known) makespan \cite{vilimfailure, van2016dynamic}.}
\begin{tabular}{c|cccccccc||c}
\toprule
TA Instances & 15$\times$15 & 20$\times$15 & 20$\times$20 & 30$\times$15 & 30$\times$20 & 50$\times$15 & 50$\times$20 & 100$\times$20 & Mean Gap\\
\midrule
ScheduleNet &\textbf{1.153}& \textbf{1.194}& \textbf{1.172} & \textbf{1.180} & \textbf{1.187} & \textbf{1.138} & \textbf{1.135} & \textbf{1.066} & \textbf{1.154}\\
\midrule
\cite{Park2021learning} & 1.171 & 1.202 & 1.228 & 1.189 & 1.254 & 1.159 & 1.174 & 1.070 & 1.181\\
\cite{zhang2020learning} & 
1.259 & 1.300 & 1.316 & 1.319 & 1.336 & 1.224 & 1.264 & 1.136 & 1.269\\
\midrule
MOR & 1.205 & 1.236 & 1.217 & 1.249 & 1.173 & 1.177 & 1.092 & 1.092 & 1.197\\
FIFO & 1.239 & 1.314 & 1.275 & 1.319 & 1.310 & 1.206 & 1.239 & 1.136 &1.255 \\
SPT & 1.258 & 1.329 & 1.278 & 1.349 & 1.344 & 1.241 & 1.256 & 1.144 & 1.275 \\
\bottomrule
\end{tabular}
\label{table:jsp-benchmark-table}
\vspace{-0.2cm}
\end{table}

\section{Conclusion}

In this work, we proposed ScheduleNet, which is a RL-based scheduler that can solve various types of multi-agent scheduling problems (mSPs) in a decentralized manner. 
We evaluated the scheduling performance on random and benchmark instances of mTSP and JSP, and verified that ScheduleNet is an effective \textit{general scheduler} that can solve various mSPs, a \textit{cooperative scheduler} that induces multi-agent coordination to achieve a common objective, and a \textit{scalable scheduler} that can solve large scale and practical scheduling problems.

\newpage

\bibliography{references}

\newpage
\appendix
\section{Details of Type-aware Graph Attention}
\label{appendix:TGA}

In this section, we thoroughly describe the computation procedures of 
type-aware graph attention (TGA). Similar to the main body, We overload notations for the simplicity of notation such that the input node and edge feature as $h_i$ and $h_{ij}$, and the embedded node and edge feature $h'_i$ and $h'_{ij}$, respectively. 

The proposed TGA performs graph embedding with the following three phases: (1) type-aware edge update, (2) type-aware message aggregation, and (3) type-aware node update. 

\textbf{Type-aware edge update.  } The edge update scheme is designed to reflect the complex type relationship among the entities while updating edge features. First, the \textit{context} embedding $c_{ij}$ of edge $e_{ij}$ is computed using the source node type $k_j$ such that: 
\begin{align}
\label{eqn:appendix-gnn-edge-context}
\begin{split}
c_{ij} &= \text{MLP}_{etype}(k_j) \\
\end{split}
\end{align}
where $\text{MLP}_{etype}$ is the edge type encoder. The source node types are embedded into the context embedding $c_{ij}$ using $\text{MLP}_{etype}$. Next, the type-aware edge encoding $u_{ij}$ is computed using the Multiplicative Interaction (MI) layer \citep{jayakumar2019multiplicative} as follows:
\begin{align}
\label{eqn:appendix-gnn-edge-mi}
\begin{split}
u_{ij} &= \text{MI}_{edge}([h_i, h_j, h_{ij}], c_{ij})\\
\end{split}
\end{align}
where $\text{MI}_{edge}$ is the edge MI layer. We utilize the MI layer, which dynamically generates its parameter depending on the context $c_{ij}$ and produces type-aware edge encoding $u_{ij}$, to effectively model the complex type relationships among the nodes. Type-aware edge encoding $u_{ij}$ can be seen as a dynamic edge feature which varies depending on the source node type. Then, the updated edge embedding $h'_{ij}$ and its attention logit $z_{ij}$ are obtained as:
\begin{align}
h'_{ij} &= \text{MLP}_{edge}(u_{ij}) \label{eqn:appendix-gnn-edge-update}\\
z_{ij} &= \text{MLP}_{attn}(u_{ij}) \label{eqn:appendix-gnn-edge-attn}
\end{align}

where $\text{MLP}_{edge}$ and $\text{MLP}_{attn}$ is the edge updater and logit function, respectively. The edge updater and logit function produce updated edge embedding and logits from the type-aware edge.

The computation steps of Equation \ref{eqn:appendix-gnn-edge-context}, \ref{eqn:appendix-gnn-edge-mi}, and \ref{eqn:appendix-gnn-edge-update} are defined as $\text{TGA}_{\mathbb{E}}$. Similarly, the computation steps of Equation \ref{eqn:appendix-gnn-edge-context}, \ref{eqn:appendix-gnn-edge-mi}, and \ref{eqn:appendix-gnn-edge-attn} are defined as $\text{TGA}_{\mathbb{A}}$.

\textbf{Type-aware message aggregation.  } We first define the type-$k$ neighborhood of $v_{i}$ as $\mathcal{N}_k(i) = \{v_l | k_l = k, \forall v_l \in \mathcal{N}(i)\}$, where $\mathcal{N}(i)$ is the in-neigborhood set of $v_{i}$. The proposed type-aware message aggregation procedure computes attention score $\alpha_{ij}$ for the $e_{ij}$, which starts from $v_{j}$ and heads to $v_{i}$, such that: 
\begin{align}
\label{eqn:tga-attention}
\begin{split}
\alpha_{ij} &= \frac{\text{exp}(z_{ij})}{\sum_{l \in \mathcal{N}_{k_j}(i)}\text{exp}(z_{il})}  \\
\end{split}
\end{align}
Intuitively speaking, The proposed attention scheme normalizes the attention logits of incoming edges over the types. Therefore, the attention scores sum up to 1 over each type-$k$ neighborhood. Next, the type-$k$ neighborhood message $m_{i}^{k}$ for node $v_{i}$ is computed as:
\begin{align}
\label{eqn:tga-intra-aggr}
\begin{split}
m_i^k = \sum_{j \in \mathcal{N}_k(i)}\alpha_{ij} h'_{ij}
\end{split}
\end{align}
In this aggregation step, the incoming messages of node $i$ are aggregated per type. All incoming type neighborhood messages are concatenated to produce (inter-type) aggregated message $m_i$ for $v_{i}$, such that:
\begin{align}
\label{eqn:tga-inter-aggr}
\begin{split}
m_i &= \text{concat}(\{m_i^k | k \in \displaystyle \sK\}) \\
\end{split}
\end{align}

\textbf{Type-aware node update.  } Similar to the edge update phase, the context embedding $c_i$ is computed first for each node $v_{i}$:
\begin{align}
\label{eqn:appendix-node-mi}
\begin{split}
c_{i} &= \text{MLP}_{ntype}(k_i)
\end{split}
\end{align}
where $\text{MLP}_{ntype}$ is the node type encoder. 
Then, the updated hidden node embedding $h'_{i}$ is computed as below:
\begin{align}
\label{eqn:appendix-node-update}
\begin{split}
h'_{i} &= \text{MLP}_{node}(h_i, u_i)\\
\end{split}
\end{align}
where $u_{i}=\text{MI}_{node}(m_i, c_i)$ is the type-aware node embedding that is produced by $\text{MI}_{node}$ layer using aggregated messages $m_i$ and the context embedding $c_i$. 

The computation steps of Equation \ref{eqn:appendix-node-mi}, and \ref{eqn:appendix-node-update} are defined as $\text{TGA}_{\mathbb{E}}$. 


\section{Extended discussion for reward normalization scheme}
In this section, we further discuss the effect of the proposed reward normalization scheme and its variants to the performance of ScheduleNet. The proposed reward normalization (i.e. normalized makespan) $m(\pi, \pi_{b})$ is given as follows:
\begin{align}
\label{appendix:eqn-reward-norm}
\begin{split}
m(\pi, \pi_{b}) =  \frac{M(\pi_{\theta})-M(\pi_b)}{M(\pi_{b})}\
\end{split}
\end{align}
where $\pi_b$ is the baseline policy. 

\textbf{Effect of the denominator. } $m(\pi, \pi_{b})$ measures the relative scheduling supremacy of $\pi$ to the $\pi_{b}$. Similar reward normalization scheme, but without $M(\pi_b)$ division, is employed to solve single-agent scheduling problems \cite{kool2018attention}. We empirically found that the division leads in much stable learning when the scale of makespan change (e.g. the area of map change from the unit square to different geometries).

\textbf{Effect of the baseline selection. } A proper selection of $\pi_b$ is essential to assure stable and asymptotically better learning of ScheduleNet. Intuitively speaking, choosing too strong baseline (i.e. policy having smaller makespan such as OR-tools) can makes the entire learning process unstable since the normalized reward tends to have larger values. On the other hand, employing too weak baseline can leads in virtually no learning since the $m(\pi, \pi_b)$ becomes nearly zero.

We select $\pi_b$ as $\text{Greedy}(\pi)$. This baseline selection has several advantages from selecting a fixed/pre-existing scheduling policy: (1) Entire learning process becomes independent from existing scheduling methods. Thus ScheduleNet is applicable even when the cheap-and-performing $\pi_b$ for some target mSP does not exist. (2) $\text{Greedy}(\pi)$ serves as an appropriate $\pi_b$ (either not too strong or weak) during policy learning. We experimentally confirmed that the baseline section $\text{Greedy}(\pi)$ results in a better scheduling policy as similar to the several literature \cite{kool2018attention, silver2016mastering}. 




\section{Experiments}

\subsection{mTSP experiments}

\subsubsection{semi-MDP formulation}
\label{appendix:mtsp-mdp-formulation}

The formulated mTSP semi-MDP is event-based. Here we discuss the further details about the event-based transitions of mTSP MDP. Whenever all agents are assigned to cities, the environment transits in time, until any of the workers arrives to the city (i.e. completes the task). Arrival of the worker to the city triggers an event, meanwhile the other assigned salesman are still on the way to their correspondingly assigned cities. We assume that each worker transits towards the assigned city with unit speed in the 2D Euclidean space, i.e. the distance travelled by each worker equals the time past between two consecutive MDP events. 

It is noteworthy that multiple events can happen at the same time, typically when time stamp $t=0$. If the MDP has multiple available workers at the same time, we repeatedly choose an arbitrary idle agent and assign it to the one of an idle task until no agent is idle, while updating event index $\tau$. This random selections do not alter the resulting solutions since we do not differentiate each agent (i.e. agents are homogeneous agents). 

\subsubsection{Agent-task graph formulation}
\label{appendix:mtsp-graph-formulation}

In this section, we present the list of all possible node types in $\gG_\tau$: (1) assigned-agent, (2) unassigned-agent, (3), (4) assigned-task, (5) unassigned-task, (5) depot. Here, we do not include already visited cities (i.e. inactive tasks) to the graph. Thus, the set of active agents/tasks is defined by the union of assigned and unassigned agents/tasks. Our empirical experiments showed no performance difference between the graph with inactive nodes included versus the graph with active-only nodes. All nodes in the $\gG_\tau$ are fully connected.

\subsubsection{Training details}
\label{appendix:mtsp-training}
\paragraph{Network parameters.}
{\fontfamily{lmtt}\selectfont ScheduleNet} is composed of two TGA layers. Each TGA layer (\textit{raw-to-hidden} and \textit{hidden-to-hidden}) has same MLP parameters, as shown in Table \ref{table-tga-parameters}. The node and edge encoders input dimensions for the first \textit{raw-to-hidden} TGA layer is 4 and 7, respectively, the output node, edge dimensions of the first TGA layer is 32, 32, which is used as input dimensions for the \textit{hidden-to-hidden} TGA layer. The embedded node and edge features are used to calculate the action embeddings via $\text{MLP}_{actor}$ with parameters described in Table \ref{table-tga-parameters}.

\begin{table*}[t]
\centering
\small\addtolength{\tabcolsep}{-1pt}
\renewcommand{\arraystretch}{1.3} 
\caption{\textbf{ScheduleNet TGA Layer parameters.}}
\vskip 0.075in
\begin{tabular}{>{\centering\arraybackslash}m{1.5cm}|>{\centering\arraybackslash}m{2cm}|>{\centering\arraybackslash}m{1.5cm}|>{\centering\arraybackslash}m{1.5cm}}
\toprule
Encoder name & MLP hidden dimensions & Hidden activation & Output activation \\
\midrule
$\text{MLP}_{etype}$ & {[}32{]}   & ReLU &   Identity    \\
$\text{MLP}_{edge}$ & {[}32, 32{]}   & ReLU   &  Identity \\
$\text{MLP}_{attn}$ & {[}32, 32{]}   & ReLU & Identity    \\
$\text{MLP}_{ntype}$ & {[}32{]}   & ReLU & Identity \\
$\text{MLP}_{node}$ & {[}32, 32{]}   & ReLU & Identity \\
$\text{MLP}_{actor}$ & {[}256, 128{]}   & ReLU & Identity\\
\bottomrule
\end{tabular}
\label{table-tga-parameters}
\end{table*}

\paragraph{Training.}
In this section, we presents a pseudocode for training ScheduleNet. We smooth the evaluation policy $\pi_\theta$ with the Polyak average as studied \cite{izmailov2018averaging} for the further stabilization of training process.

\begin{algorithm}[H]
\DontPrintSemicolon
\caption{ScheduleNet Training}
\SetKwInOut{Input}{input}
\SetKwInOut{Output}{output}
\Input {Training policy $\pi_\theta$}
\Output {Smoothed policy $\pi_\phi$}

Initialize the smoothed policy with parameters $\phi \leftarrow \theta$. \\
\For{$\text{update step}$}
{Generate a random mTSP instance $I$ \\
$\pi_b \leftarrow \text{Greedy}(\pi_\theta)$ \\
\For{$\text{number of episodes}$}
    {Construct mTSP MDP from the instance $I$ \\
    Collect samples with $\pi_\theta$ and $\pi_b$ from the mTSP MDP.
    }
\For{$\text{inner updates K}$}
    {
    $\theta \leftarrow \theta + \alpha \nabla_\theta \mathcal{L}(\theta)$
    }
$\phi \leftarrow \beta \phi + (1-\beta) \theta$
}
\caption{ScheduleNet Training}
\end{algorithm}
We set the MDP discounting factor $\gamma$ to 0.9, and the polyak smoothing coefficient $\beta$ as 0.1, and the clipping parameter $\epsilon$ to 0.2.

\subsubsection{Details of Random mTSP experiments}
\label{appendix:mtsp-random-results}
\paragraph{OR-Tools implementation.} The Google \texttt{OR-Tools} routing module \cite{ortools} is a set of highly optimized and also practically working meta-heuristics for solving various routing problems (e.g. mTPS, mVRP). It first finds the initial feasible solution and then iteratively improves the solution with local search heuristics (e.g. Greedy Descent, Simulated Annealing, Tabu Search) until certain termination condition is satisfied. We acheive the scheduling results of the baseline OR-tools algorithm with the official implementation mTSP provided by Google. 

We further tune the hyperparameters of \texttt{OR-Tools} to achieve better scheduling results on large mTSP instances ($n >200$ and $m >20$) by applying different initial solution search and solution improvement schemes. However, such tuning results in virtually no improvements in scheduling performances.

\paragraph{Two-phase heuristics implementations.}
The 2-phase heuristics for mTSP is an extension of well-known TSP heuristics to the $m>1$ cases. First, we perform $K$-means clustering (where $K=m$) of cities in the mTSP instance by utilizing \texttt{scikit-learn} \cite{scikit-learn}. Next, we apply TSP insertion heuristics, Nearest Insertion, Farthest Insertion, Random Insertion, and Nearest Neighbour Insertion, for each cluster of cities. It should be noted that, performance of the 2-phase heuristics is highly depended on the spatial distribution of the cities on the map. Thus 2-phase heuristics perform particularly well on uniformly distributed random instances, where $K$-means clustering can obtain clusters with approximately same number of cities per cluster.

\subsubsection{Qualitative analysis}
\begin{figure}[t]
\begin{center}
\includegraphics[width={\linewidth}]{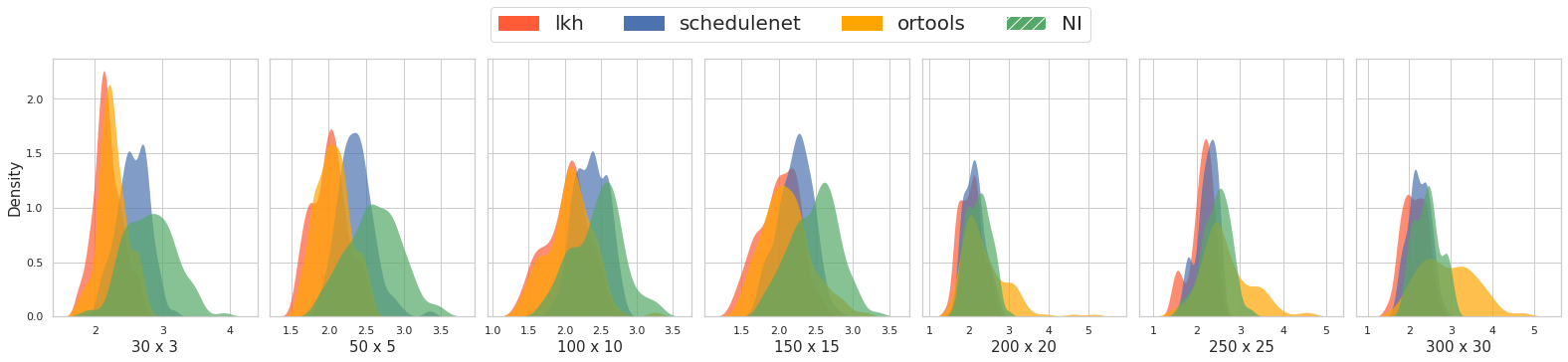}
\end{center}
\caption{\textbf{Histogram of makespans on random ($N\times m$) mTSP datasets.} The x-axis demonstrates the makespans. The y-axis shows the density of makespans.}
\label{fig:appendix-mtsp-random-histogram}
\end{figure}


\paragraph{The histogram of scheduling performances on random mTSP datasets.} ScheduleNet averagely shows higher makespans than OR-tools on the small ($M<200$) random mTSP datasets as discussed in Section \ref{section:mtsp-experiments}. However, we observe that, even on the small mTSP problems, ScheduleNet can outperform OR-tools as visualized in Figure \ref{fig:appendix-mtsp-random-histogram}.



\begin{figure}[t]
\begin{center}
\includegraphics[width={0.5\linewidth}]{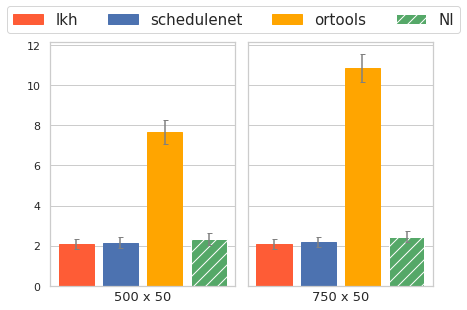}
\end{center}
\caption{\textbf{Scheduling performances on large random ($N\times m$) mTSP datasets.}  The y-axis shows the makespans.}
\label{fig:appendix-mtsp-random-knn-xl}
\end{figure}

\paragraph{Extended results of sparse graph experiments.} We provide the scheduling performance of ScheduleNet and baseline algorithms on random mTSPs of size ($500 \times 50$) and ($750 \times 70$). See Figure \ref{fig:appendix-mtsp-random-knn-xl} for the results.

\subsection{JSP experiments}
\begin{figure}[t]
  \begin{minipage}{0.55\textwidth}
     \centering
     \includegraphics[width=.99\linewidth]{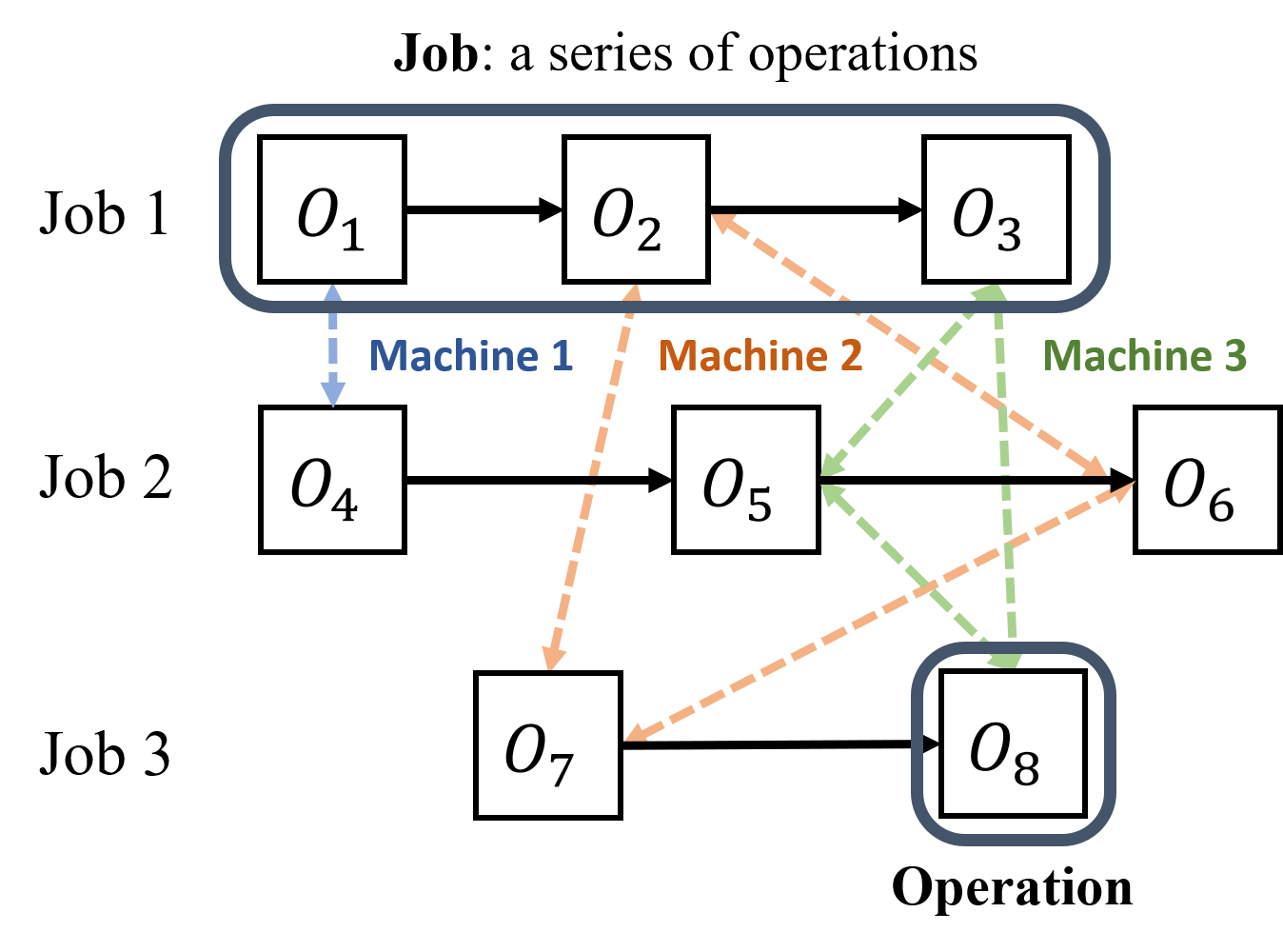}
     \caption{\textbf{Disjunctive graph representation of JSP}}
     \label{fig:JSP-disjunctive-repr}
  \end{minipage} 
  \hfill
  \begin{minipage}{0.39\textwidth}
     \centering
     \vskip 0.175in
     \includegraphics[width=.99\linewidth]{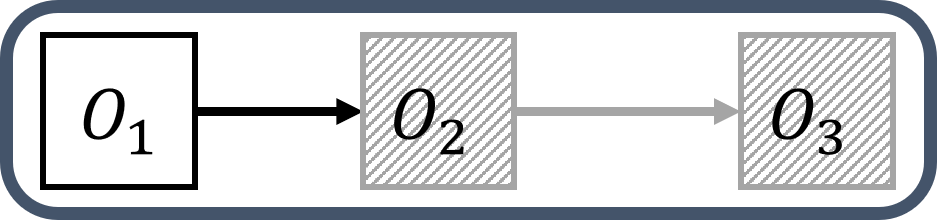}
     \caption{\textbf{Precedence constraint}}
     \label{fig:JSP-precedence-const}
     \vskip 0.150in
     \centering
     \includegraphics[width=.8\linewidth]{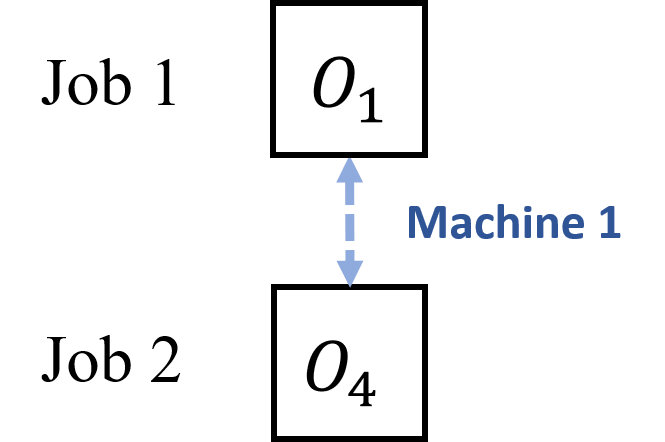}
     \caption{\textbf{Agent-sharing constraint}}
     \label{fig:JSP-disjunctive-const}
  \end{minipage}
\end{figure}

Job-shop scheduling problem (JSP) is a mSP that can be applied in various industries including the operation of semi-conductor chip fabrication facility and railroad system. The objective of JSP is to find the sequence of machine (agent) allocations to finish the jobs (a sequence of operations; tasks) as soon as possible. JSP can be seen as an extension of mTSP with two additional constraints: (1) precedence constraint that models a scenario where an operation of a job becomes processable only after all of its preceding operations are done; (2) agent-sharing (disjunctive) constraint that confines the machine to process only one operation at a time. Due to these additional constraints, JSP is considered to be a more difficult problem when it is solved via mathematical optimization techniques. A common representation of JSP is the disjunctive graph representation. As shown in Figures \ref{fig:JSP-disjunctive-repr}, \ref{fig:JSP-precedence-const} and \ref{fig:JSP-disjunctive-const}, JSP contains the set of jobs, machines, precedence constraints, and disjunctive constraints as its entities. In the following subsections, we provide the details of the proposed MDP formulation of JSP, training details of ScheduleNet, and experiment results.

\subsubsection{semi-MDP formulation}
\label{appendix:jsp-mdp-formulation}
The semi-MDP formulation of JSP is similar to that of mTSP. The specific definitions of the state and action for JSP are as follows:

\textbf{State.} We define $s_{\tau}=(\{s_{\tau}^{i}\}_{i=1}^{N+m}, s_{\tau}^{\text{env}})$ which is composed of two types of states: entity state $s_{\tau}^{i}$ and environment state $s_{\tau}^{\text{env}}$.
\vspace{-0.35cm}
\begin{itemize}[leftmargin=*]
\setlength\itemsep{-0.1em}
    \item $s_{\tau}^{i} = (p_{\tau}^{i}, \1_{\tau}^{\text{processable}},\1_{\tau}^{\text{assigned}}$,  $\1_{\tau}^{\text{accessible}}$,  $\1_{\tau}^{\text{waiting}})$ is the state of the $i$-th entity. $p_{\tau}^{i}$ is the processing time of the $i$-th entity at the $\tau$-th event. $\1_{\tau}^{\text{processable}}$ indicates whether the $i$-th task is processable by the target agent or not. Similar to mTSP, $\1_{\tau}^{\text{assigned}}$ indicates whether an agent/task is assigned.
    \item $s_{\tau}^{\text{env}}$ contains the current time of the environment, the sequence of tasks completed by each agent (machine), and the precedence constraints of tasks within each job.
\end{itemize}

\textbf{Action.} We define the action space at the $\tau$-th event as a set of oeprations that is both processable and currently available. Additionally, we define the \textit{waiting} action as a reservation of the target agent (i.e. the unique idle machine) until the next event. Having \textit{waiting} as an action allows the adaptive scheduler (e.g. ScheduleNet) to achieve the optimal scheduling solution (and also makespan) from the JSP MDP, where the optimal solution contains waiting (idle) time intervals.

\subsubsection{Agent-task graph formulation}
\label{appendix:jsp-graph-formulation}

ScheduleNet constructs the \textit{agent-task graph} $\gG_\tau$ that reflects the complex relationships among the entities in $s_\tau$.
ScheduleNet constructs a directed graph $\gG_\tau = (\sV, \mathbb{E})$ out of $s_{\tau}$, where $\sV$ is the set of nodes and $\mathbb{E}$ is the set of edges. The nodes and edges and their associated features are defined as:
\begin{itemize}[leftmargin=*]
\setlength\itemsep{-0.1em}
    \item $v_i$ denotes the $i$-th node and represents either an agent or a task. $v_i$ contains the node feature $x_i=(s_{\tau}^i,k_i)$, where $s_{\tau}^i$ is the state of entity $i$, and $k_i$ is the type of $v_i$ (e.g. if the entity $i$ is a \textit{task} and its $\1_\tau^{processable}=1$, then the $k_i$ becomes a \textit{processable-task} type.)
    \item $e_{ij}$ denotes the edge between the source node $v_j$ and the destination node $v_i$. The edge feature $w_{ij}$ is a binary feature which indicates whether the destination node $v_i$ is processable by the source node $v_j$.
\end{itemize}
All possible node types in $\gG_\tau$ are: (1) assigned-agent, (2) unassigned-agent, (3) assigned-task, (4) processable-task, and (5) unprocessable-task. We do not include completed tasks in the graph. Thus, the currently active tasks are the union of the assigned tasks, processable-tasks, and unprocessable-tasks. The full list of node features are as follows:
\begin{itemize}[leftmargin=*]
\setlength\itemsep{-0.1em}
    \item $\1_{\tau}^{\text{agent}}$ indicates whether the node is a agent or a task.
    \item $\1_{\tau}^{\text{target-agent}}$ indicates whether the node is a target-agent (unique idle agent that needs to be assigned).
    \item $\1_{\tau}^{\text{assigned}}$ indicates whether the agent/task is assigned.
    \item $\1_{\tau}^{\text{waiting}}$ indicates whether the node is an agent in \textit{waiting} state.
    \item $\1_{\tau}^{\text{processable}}$ indicates whether the node is a task that is \textit{processable} by the target-agent.
    \item $\1_{\tau}^{\text{accessible}}$ indicates whether the node is processable by the target-agent and is available.
    \item \textit{Task wait time} indicates the amount of time passed since the operation became \textit{accessible}.
    \item \textit{Task processing time} indicates the processing time of the operation.
    \item \textit{Time-to-complete} indicates the amount of time it will take to complete the task, i.e. the time-distance to the given task.
    \item \textit{Remain ops.} indicates the number of remaining operations to be completed for the job where the task belongs to.
    \item \textit{Job completion ratio} is the ratio of completed operations within the job to the total amount of operations in the job.
\end{itemize}

\begin{figure}[t]
\begin{center}
\includegraphics[width={0.70\linewidth}]{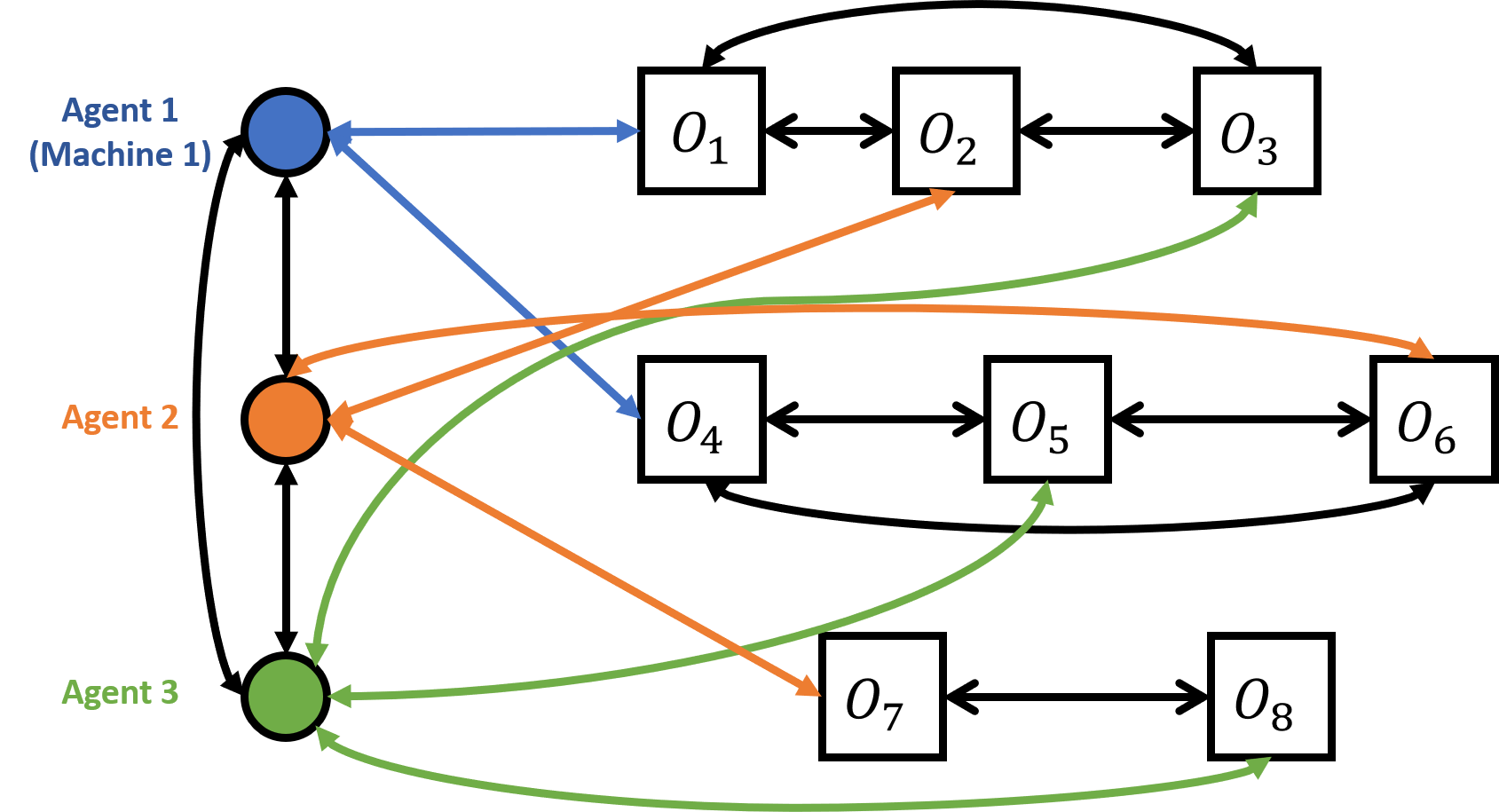}
\end{center}
\caption{\textbf{JSP agent-task graph representation}}
\label{fig:JSP-graph-repr}
\end{figure}

\textbf{JSP graph connectivity}. Figure \ref{fig:JSP-graph-repr} visualizes the proposed agent-task graph. From Figure \ref{fig:JSP-graph-repr}, each agent is fully connected to the set of processable tasks by that agent, and vice versa. Each task is fully connected to the other tasks (operations) that belong to the same job. Each agent is fully connected to the other agents. 

\subsubsection{Training details}
The TGA layer hyperparameters are set according to Table \ref{table-tga-parameters}. We use same training hyperparameters as in Appendix \ref{appendix:mtsp-training}.
The node and edge encoder input dimensions for the first \textit{raw-to-hidden} TGA layer is 12 and 4 respectively. The output node and edge dimensions of the first TGA layer is 32 and 32 respectively, which are used as the input dimensions for the \textit{hidden-to-hidden} TGA layer.

\label{appendix:jsp-training}
\subsubsection{Random JSP experiments}

\paragraph{CP-SAT implementation.} 
CP-SAT is one of the state-of-the-art constraint programming solver that is implemented as a module of Google \texttt{OR-Tools} \cite{ortools}. We employ CP-SAT with a one hour time-limit as a baseline algorithm for solving JSP. Our implementation of CP-SAT is directly adopted from the official JSP solver implementation provided by Google, which is considered to be highly optimized to solve JSP. The official implementation can be found in \url{https://developers.google.com/optimization/scheduling/job_shop}.

\paragraph{JSP heuristic implementations.} Priority dispatching rules (PDR) is one of the most common JSP solving heuristics. PDR computes the priority of the feasible operations (i.e. the set of operations whose precedent operation is done and, at the same time, the target machine is idle) by utilizing the dispatching rules. As the JSP heuristic baselines, we consider the following three dispatching rules:
\begin{itemize}[leftmargin=*]
\setlength\itemsep{-0.1em}
    \item Most Operation Remaining (MOR) sets the highest priority to the operation that has the most remaining operations to finish its corresponding job.
    \item First-in-first-out (FIFO) sets the highest priority to the operation that joins to the feasible operation set first.
    \item Shortest Processing Time (SPT) sets the highest priority to the operation that has the shortest processing time.
\end{itemize}

\subsubsection{Extended public benchmark JSP results}
\label{appendix:jsp-benchmark-results}
We provide the detailed JSP results for the following public datasets:
TA \citep{taillard1993benchmarks} (Tables \ref{table-ta1} and \ref{table-ta2}),
ORB \cite{applegate1991computational}, FT \cite{fisher1963probabilistic}, YN \cite{yamada1992genetic} (Table \ref{table-abz}), SWV \cite{storer1992new} (Table \ref{table-SWV}), and LA \cite{lawrence1984resouce} (Table \ref{table-LA}).

\begin{table*}[t]
\centering
\renewcommand{\arraystretch}{1.3} 
\caption{\textbf{Job-shop scheduling makespans on TA dataset (Part 1)}}
\vskip 0.05in
\resizebox{\textwidth}{!}{
\begin{tabular}{c|c|ccc|cc|c|c}
\toprule
\textbf{Instance} & $\mathbf{N} \times \mathbf{m}$   & \textbf{SPT}  & \textbf{FIFO} & \textbf{MOR} & \citet{Park2021learning} & \citet{zhang2020learning} & \textbf{ScheduleNet} & \textbf{OPT}  \\
\midrule
Ta01     & $15 \times 15$ & 1462 & 1486 & 1438 & 1389 & 1443    & 1452 & 1231 \\
Ta02     & $15 \times 15$ & 1446 & 1486 & 1452 & 1519 & 1544    & 1411 & 1244 \\
Ta03     & $15 \times 15$ & 1495 & 1461 & 1418 & 1457 & 1440    & 1396 & 1218 \\
Ta04     & $15 \times 15$ & 1708 & 1575 & 1457 & 1465 & 1637    & 1348 & 1175 \\
Ta05     & $15 \times 15$ & 1618 & 1457 & 1448 & 1352 & 1619    & 1382 & 1224 \\
Ta06     & $15 \times 15$ & 1522 & 1528 & 1486 & 1481 & 1601    & 1413 & 1238 \\
Ta07     & $15 \times 15$ & 1434 & 1497 & 1456 & 1554 & 1568    & 1380 & 1227 \\
Ta08     & $15 \times 15$ & 1457 & 1496 & 1482 & 1488 & 1468    & 1374 & 1217 \\
Ta09     & $15 \times 15$ & 1622 & 1642 & 1594 & 1556 & 1627    & 1523 & 1274 \\
Ta10     & $15 \times 15$ & 1697 & 1600 & 1582 & 1501 & 1527    & 1493 & 1241 \\
\midrule
Ta11     & $20 \times 15$ & 1865 & 1701 & 1665 & 1626 & 1794    & 1612 & 1357 \\
Ta12     & $20 \times 15$ & 1667 & 1670 & 1739 & 1668 & 1805    & 1600 & 1367 \\
Ta13     & $20 \times 15$ & 1802 & 1862 & 1642 & 1715 & 1932    & 1625 & 1342 \\
Ta14     & $20 \times 15$ & 1635 & 1812 & 1662 & 1642 & 1664    & 1590 & 1345 \\
Ta15     & $20 \times 15$ & 1835 & 1788 & 1682 & 1672 & 1730    & 1676 & 1339 \\
Ta16     & $20 \times 15$ & 1965 & 1825 & 1638 & 1700 & 1710    & 1550 & 1360 \\
Ta17     & $20 \times 15$ & 2059 & 1899 & 1856 & 1678 & 1897    & 1753 & 1462 \\
Ta18     & $20 \times 15$ & 1808 & 1833 & 1710 & 1684 & 1794    & 1668 & 1396 \\
Ta19     & $20 \times 15$ & 1789 & 1716 & 1651 & 1900 & 1682    & 1622 & 1332 \\
Ta20     & $20 \times 15$ & 1710 & 1827 & 1622 & 1752 & 1739    & 1604 & 1348 \\
\midrule
Ta21     & $20 \times 20$ & 2175 & 2089 & 1964 & 2199 & 2252    & 1921 & 1642 \\
Ta22     & $20 \times 20$ & 1965 & 2146 & 1905 & 2049 & 2102    & 1844 & 1600 \\
Ta23     & $20 \times 20$ & 1933 & 2010 & 1922 & 2006 & 2085    & 1879 & 1557 \\
Ta24     & $20 \times 20$ & 2230 & 1989 & 1943 & 2020 & 2200    & 1922 & 1644 \\
Ta25     & $20 \times 20$ & 1950 & 2160 & 1957 & 1981 & 2201    & 1897 & 1595 \\
Ta26     & $20 \times 20$ & 2188 & 2182 & 1964 & 2057 & 2176    & 1887 & 1643 \\
Ta27     & $20 \times 20$ & 2096 & 2091 & 2160 & 2187 & 2132    & 2009 & 1680 \\
Ta28     & $20 \times 20$ & 1968 & 1980 & 1952 & 2054 & 2146    & 1813 & 1603 \\
Ta29     & $20 \times 20$ & 2166 & 2011 & 1899 & 2210 & 1952    & 1875 & 1625 \\
Ta30     & $20 \times 20$ & 1999 & 1941 & 2017 & 2140 & 2035    & 1913 & 1584 \\
\midrule
Ta31     & $30 \times 15$ & 2335 & 2277 & 2143 & 2251 & 2565    & 2055 & 1764 \\
Ta32     & $30 \times 15$ & 2432 & 2279 & 2188 & 2378 & 2388    & 2268 & 1784 \\
Ta33     & $30 \times 15$ & 2453 & 2481 & 2308 & 2316 & 2324    & 2281 & 1791 \\
Ta34     & $30 \times 15$ & 2434 & 2546 & 2193 & 2319 & 2332    & 2061 & 1829 \\
Ta35     & $30 \times 15$ & 2497 & 2478 & 2255 & 2333 & 2505    & 2218 & 2007 \\
Ta36     & $30 \times 15$ & 2445 & 2433 & 2307 & 2210 & 2497    & 2154 & 1819 \\
Ta37     & $30 \times 15$ & 2664 & 2382 & 2190 & 2201 & 2325    & 2112 & 1771 \\
Ta38     & $30 \times 15$ & 2155 & 2277 & 2179 & 2151 & 2302    & 1970 & 1673 \\
Ta39     & $30 \times 15$ & 2477 & 2255 & 2167 & 2138 & 2410    & 2146 & 1795 \\
Ta40     & $30 \times 15$ & 2301 & 2069 & 2028 & 2007 & 2140    & 2030 & 1669 \\
\bottomrule
\end{tabular}
}
\label{table-ta1}
\vspace{-0.2cm}
\end{table*}
\begin{table*}[t]
\centering
\renewcommand{\arraystretch}{1.3}
\caption{\textbf{Job-shop scheduling makespans on TA dataset (Part 2)}}
\vskip 0.05in
\resizebox{\textwidth}{!}{
\begin{tabular}{c|c|ccc|cc|c|c}
\toprule
\textbf{Instance} & $\mathbf{N} \times \mathbf{m}$   & \textbf{SPT}  & \textbf{FIFO} & \textbf{MOR} & \citet{Park2021learning} & \citet{zhang2020learning} & \textbf{ScheduleNet} & \textbf{OPT}  \\
\midrule
Ta41     & $30 \times 20$ & 2499 & 2543 & 2538 & 2654 & 2667    & 2572 & 2005 \\
Ta42     & $30 \times 20$ & 2710 & 2669 & 2440 & 2579 & 2664    & 2397 & 1937 \\
Ta43     & $30 \times 20$ & 2434 & 2506 & 2432 & 2737 & 2431    & 2310 & 1846 \\
Ta44     & $30 \times 20$ & 2906 & 2540 & 2426 & 2772 & 2714    & 2456 & 1979 \\
Ta45     & $30 \times 20$ & 2640 & 2565 & 2487 & 2435 & 2637    & 2445 & 2000 \\
Ta46     & $30 \times 20$ & 2667 & 2582 & 2490 & 2681 & 2776    & 2541 & 2004 \\
Ta47     & $30 \times 20$ & 2620 & 2508 & 2286 & 2428 & 2476    & 2280 & 1889 \\
Ta48     & $30 \times 20$ & 2620 & 2541 & 2371 & 2440 & 2490    & 2358 & 1941 \\
Ta49     & $30 \times 20$ & 2666 & 2550 & 2397 & 2446 & 2556    & 2301 & 1961 \\
Ta50     & $30 \times 20$ & 2429 & 2531 & 2469 & 2530 & 2628    & 2453 & 1923 \\
\midrule
Ta51     & $50 \times 15$ & 3856 & 3590 & 3567 & 3145 & 3599    & 3382 & 2760 \\
Ta52     & $50 \times 15$ & 3266 & 3365 & 3303 & 3157 & 3341    & 3231 & 2756 \\
Ta53     & $50 \times 15$ & 3507 & 3169 & 3115 & 3103 & 3186    & 3083 & 2717 \\
Ta54     & $50 \times 15$ & 3142 & 3218 & 3265 & 3278 & 3266    & 3068 & 2839 \\
Ta55     & $50 \times 15$ & 3225 & 3291 & 3279 & 3142 & 3232    & 3078 & 2679 \\
Ta56     & $50 \times 15$ & 3530 & 3329 & 3100 & 3258 & 3378    & 3065 & 2781 \\
Ta57     & $50 \times 15$ & 3725 & 3654 & 3335 & 3230 & 3471    & 3266 & 2943 \\
Ta58     & $50 \times 15$ & 3365 & 3362 & 3420 & 3469 & 3732    & 3321 & 2885 \\
Ta59     & $50 \times 15$ & 3294 & 3357 & 3117 & 3108 & 3381    & 3044 & 2655 \\
Ta60     & $50 \times 15$ & 3500 & 3129 & 3044 & 3256 & 3352    & 3036 & 2723 \\
\midrule
Ta61     & $50 \times 20$ & 3606 & 3690 & 3376 & 3425 & 3654    & 3202 & 2868 \\
Ta62     & $50 \times 20$ & 3639 & 3657 & 3417 & 3626 & 3722    & 3339 & 2869 \\
Ta63     & $50 \times 20$ & 3521 & 3367 & 3276 & 3110 & 3536    & 3118 & 2755 \\
Ta64     & $50 \times 20$ & 3447 & 3179 & 3057 & 3329 & 3631    & 2989 & 2702 \\
Ta65     & $50 \times 20$ & 3332 & 3273 & 3249 & 3339 & 3359    & 3168 & 2725 \\
Ta66     & $50 \times 20$ & 3677 & 3610 & 3335 & 3340 & 3555    & 3199 & 2845 \\
Ta67     & $50 \times 20$ & 3487 & 3612 & 3392 & 3371 & 3567    & 3236 & 2825 \\
Ta68     & $50 \times 20$ & 3336 & 3471 & 3251 & 3265 & 3680    & 3072 & 2784 \\
Ta69     & $50 \times 20$ & 3862 & 3607 & 3526 & 3798 & 3592    & 3535 & 3071 \\
Ta70     & $50 \times 20$ & 3801 & 3784 & 3590 & 3919 & 3643    & 3436 & 2995 \\
\midrule
Ta71     & $100 \times 20$ & 6232 & 6270 & 5938 & 5962 & 6452    & 5879 & 5464 \\
Ta72     & $100 \times 20$ & 5973 & 5671 & 5639 & 5522 & 5695    & 5456 & 5181 \\
Ta73     & $100 \times 20$ & 6482 & 6357 & 6128 & 6335 & 6462    & 6052 & 5568 \\
Ta74     & $100 \times 20$ & 6062 & 6003 & 5642 & 5827 & 5885    & 5513 & 5339 \\
Ta75     & $100 \times 20$ & 6217 & 6420 & 6212 & 6042 & 6355    & 5992 & 5392 \\
Ta76     & $100 \times 20$ & 6370 & 6183 & 5936 & 5707 & 6135    & 5773 & 5342 \\
Ta77     & $100 \times 20$ & 6045 & 5952 & 5829 & 5737 & 6056    & 5637 & 5436 \\
Ta78     & $100 \times 20$ & 6143 & 6328 & 5886 & 5979 & 6101    & 5833 & 5394 \\
Ta79     & $100 \times 20$ & 6018 & 6003 & 5652 & 5799 & 5943    & 5556 & 5358 \\
Ta80     & $100 \times 20$ & 5848 & 5763 & 5707 & 5718 & 5892    & 5545 & 5183 \\
\bottomrule
\end{tabular}
}
\label{table-ta2}
\vspace{-0.2cm}
\end{table*}

\begin{table*}[t]
\centering
\renewcommand{\arraystretch}{1.3} 
\caption{\textbf{Job-shop scheduling makespans on ABZ, FT, ORB, and YN datasets.}}
\vskip 0.05in
\resizebox{0.9 \textwidth}{!}{
\begin{tabular}{c|c|ccc|c|c|c}
\toprule
\textbf{Instance} & $\mathbf{N} \times \mathbf{m}$   & \textbf{SPT}  & \textbf{FIFO} & \textbf{MOR} & \citet{Park2021learning} & \textbf{ScheduleNet} & \textbf{OPT}  \\
\midrule
abz5 & $10 \times 10$ & 1352 & 1467 & 1336 & 1353 & 1336 & 1234 \\
abz6 & $10 \times 10$ & 1097 & 1045 & 1031 & 1043 & 981  & 943  \\
abz7 & $20 \times 15$ & 849  & 803  & 775  & 887  & 791  & 656  \\
abz8 & $20 \times 15$ & 929  & 877  & 810  & 843  & 787  & 665  \\
abz9 & $20 \times 15$ & 887  & 946  & 899  & 848  & 832  & 678  \\
\midrule
ft06 & $6 \times 6$ & 88   & 65   & 59   & 71   & 59   & 55   \\
ft10 & $10 \times 10$ & 1074 & 1184 & 1163 & 1142 & 1111 & 930  \\
ft20 & $20 \times 5$ & 1267 & 1645 & 1601 & 1338 & 1498 & 1165 \\
\midrule
orb01 & $10 \times 10$ & 1478 & 1368 & 1307 & 1336 & 1276 & 1059 \\
orb02 & $10 \times 10$ & 1175 & 1007 & 1047 & 1067 & 958  & 888  \\
orb03 & $10 \times 10$ & 1179 & 1405 & 1445 & 1202 & 1335 & 1005 \\
orb04 & $10 \times 10$ & 1236 & 1325 & 1287 & 1281 & 1178 & 1005 \\
orb05 & $10 \times 10$ & 1152 & 1155 & 1050 & 1082 & 1042 & 887  \\
orb06 & $10 \times 10$ & 1190 & 1330 & 1345 & 1178 & 1222 & 1010 \\
orb07 & $10 \times 10$ & 504  & 475  & 500  & 477  & 456  & 397  \\
orb08 & $10 \times 10$ & 1107 & 1225 & 1278 & 1156 & 1178 & 899  \\
orb09 & $10 \times 10$ & 1262 & 1189 & 1165 & 1143 & 1145 & 934  \\
orb10 & $10 \times 10$ & 1113 & 1303 & 1256 & 1087 & 1080 & 944  \\
\midrule
yn1 & $20 \times 20$   & 1196 & 1113 & 1045 & 1118 & 1027 & 884  \\
yn2 & $20 \times 20$  & 1256 & 1148 & 1074 & 1097 & 1037 & 904  \\
yn3 & $20 \times 20$  & 1042 & 1135 & 1100 & 1083 & 1046 & 892  \\
yn4 & $20 \times 20$  & 1273 & 1194 & 1267 & 1258 & 1216 & 968  \\
\bottomrule
\end{tabular}
}
\label{table-abz}
\vspace{-0.2cm}
\end{table*}
\begin{table*}[t]
\centering
\renewcommand{\arraystretch}{1.3} 
\caption{\textbf{Job-shop scheduling makespans on SWV datasets.}}
\vskip 0.05in
\resizebox{0.9 \textwidth}{!}{
\begin{tabular}{c|c|ccc|c|c|c}
\toprule
\textbf{Instance} & $\mathbf{N} \times \mathbf{m}$   & \textbf{SPT}  & \textbf{FIFO} & \textbf{MOR} & \citet{Park2021learning} & \textbf{ScheduleNet} & \textbf{OPT}  \\
\midrule
swv01 & $20 \times 10$ & 1737 & 2154 & 1971 & 1761 & 1913 & 1407 \\
swv02 & $20 \times 10$ & 1706 & 2157 & 2158 & 1846 & 1998 & 1475 \\
swv03 & $20 \times 10$ & 1806 & 2019 & 1870 & 1892 & 1830 & 1398 \\
swv04 & $20 \times 10$ & 1874 & 2015 & 2026 & 1908 & 1971 & 1464 \\
swv05 & $20 \times 10$ & 1922 & 2003 & 2049 & 1796 & 1922 & 1424 \\
\midrule
swv06 & $20 \times 15$ & 2140 & 2519 & 2287 & 2068 & 2216 & 1671 \\
swv07 & $20 \times 15$ & 2146 & 2268 & 2101 & 2194 & 2037 & 1594 \\
swv08 & $20 \times 15$ & 2231 & 2554 & 2480 & 2191 & 2255 & 1752 \\
swv09 & $20 \times 15$ & 2247 & 2498 & 2553 & 2278 & 2196 & 1655 \\
swv10 & $20 \times 15$ & 2337 & 2352 & 2431 & 2141 & 2279 & 1743 \\
\midrule
swv11 & $50 \times 10$ & 3714 & 4427 & 4642 & 3989 & 4390 & 2983 \\
swv12 & $50 \times 10$ & 3759 & 4749 & 4821 & 4136 & 4532 & 2977 \\
swv13 & $50 \times 10$ & 3657 & 4829 & 4755 & 4008 & 4602 & 3104 \\
swv14 & $50 \times 10$ & 3506 & 4621 & 4740 & 3758 & 4387 & 2968 \\
swv15 & $50 \times 10$ & 3501 & 4620 & 4905 & 3860 & 4402 & 2885 \\
swv16 & $50 \times 10$ & 3453 & 2951 & 2924 & 2924 & 2924 & 2924 \\
swv17 & $50 \times 10$ & 3082 & 2962 & 2848 & 2840 & 2794 & 2794 \\
swv18 & $50 \times 10$ & 3191 & 2974 & 2852 & 2852 & 2852 & 2852 \\
swv19 & $50 \times 10$ & 3161 & 3095 & 3060 & 2961 & 2992 & 2843 \\
swv20 & $50 \times 10$ & 3125 & 2853 & 2851 & 2823 & 2823 & 2823 \\
\bottomrule
\end{tabular}
}
\label{table-SWV}
\vspace{-0.2cm}
\end{table*}
\begin{table*}[t]
\centering
\renewcommand{\arraystretch}{1.3} 
\caption{\textbf{Job-shop scheduling makespans on LA datasets.}}
\vskip 0.05in
\resizebox{0.9 \textwidth}{!}{
\begin{tabular}{c|c|ccc|c|c|c}
\toprule
\textbf{Instance} & $\mathbf{N} \times \mathbf{m}$   & \textbf{SPT}  & \textbf{FIFO} & \textbf{MOR} & \citet{Park2021learning} & \textbf{ScheduleNet} & \textbf{OPT}  \\
\midrule
la01 & $10 \times 5$  & 751  & 772  & 763  & 805  & 680 & 666  \\
la02 & $10 \times 5$ & 821  & 830  & 812  & 687  & 768  & 655  \\
la03 & $10 \times 5$ & 672  & 755  & 726  & 862  & 734  & 597  \\
la04 & $10 \times 5$ & 711  & 695  & 706  & 650  & 698  & 590  \\
la05 & $10 \times 5$ & 610  & 610  & 593  & 593  & 593  & 593  \\
\midrule
la06 & $15 \times 5$ & 1200 & 926  & 926  & 926  & 926  & 926  \\
la07 & $15 \times 5$ & 1034 & 1088 & 1001 & 931  & 1008 & 890  \\
la08 & $15 \times 5$ & 942  & 980  & 925  & 863  & 863  & 863  \\
la09 & $15 \times 5$ & 1045 & 1018 & 951  & 951  & 951  & 951  \\
la10 & $15 \times 5$ & 1049 & 1006 & 958  & 966  & 958  & 958  \\
\midrule
la11 & $20 \times 5$ & 1473 & 1272 & 1222 & 1276 & 1254 & 1222 \\
la12 & $20 \times 5$ & 1203 & 1039 & 1039 & 1039 & 1039 & 1039 \\
la13 & $20 \times 5$ & 1275 & 1199 & 1150 & 1150 & 1150 & 1150 \\
la14 & $20 \times 5$ & 1427 & 1292 & 1292 & 1292 & 1292 & 1292 \\
la15 & $20 \times 5$ & 1339 & 1587 & 1436 & 1282 & 1395 & 1207 \\
\midrule
la16 & $10 \times 10$ & 1156 & 1180 & 1108 & 1134 & 1047 & 945  \\
la17 & $10 \times 10$ & 924  & 943  & 844  & 953  & 888  & 784  \\
la18 & $10 \times 10$ & 981  & 1049 & 942  & 1049 & 947  & 848  \\
la19 & $10 \times 10$ & 940  & 983  & 1088 & 880  & 963  & 842  \\
la20 & $10 \times 10$ & 1000 & 1272 & 1130 & 1042 & 989  & 902  \\
\midrule
la21 & $15 \times 10$ & 1324 & 1265 & 1251 & 1309 & 1261 & 1046 \\
la22 & $15 \times 10$ & 1180 & 1312 & 1198 & 1158 & 1027 & 927  \\
la23 & $15 \times 10$ & 1162 & 1354 & 1268 & 1085 & 1145 & 1032 \\
la24 & $15 \times 10$ & 1203 & 1141 & 1149 & 1129 & 1088 & 935  \\
la25 & $15 \times 10$ & 1449 & 1283 & 1209 & 1308 & 1117 & 977  \\
\midrule
la26 & $20 \times 10$ & 1498 & 1372 & 1411 & 1553 & 1458 & 1218 \\
la27 & $20 \times 10$ & 1784 & 1644 & 1566 & 1624 & 1516 & 1235 \\
la28 & $20 \times 10$ & 1610 & 1474 & 1477 & 1438 & 1357 & 1216 \\
la29 & $20 \times 10$ & 1556 & 1540 & 1437 & 1582 & 1320 & 1152 \\
la30 & $20 \times 10$ & 1792 & 1648 & 1565 & 1649 & 1490 & 1355 \\
\midrule
la31 & $30 \times 10$ & 1951 & 1918 & 1836 & 1817 & 1906 & 1784 \\
la32 & $30 \times 10$ & 2165 & 2110 & 1984 & 1977 & 1850 & 1850 \\
la33 & $30 \times 10$ & 1901 & 1873 & 1811 & 1795 & 1731 & 1719 \\
la34 & $30 \times 10$ & 2070 & 1925 & 1853 & 1895 & 1784 & 1721 \\
la35 & $30 \times 10$ & 2118 & 2142 & 2064 & 2041 & 1969 & 1888 \\
\midrule
la36 & $15 \times 15$ & 1799 & 1516 & 1492 & 1489 & 1449 & 1268 \\
la37 & $15 \times 15$ & 1655 & 1873 & 1606 & 1623 & 1653 & 1397 \\
la38 & $15 \times 15$ & 1404 & 1475 & 1455 & 1421 & 1444 & 1196 \\
la39 & $15 \times 15$ & 1534 & 1532 & 1540 & 1555 & 1430 & 1233 \\
la40 & $15 \times 15$ & 1476 & 1531 & 1358 & 1570 & 1357 & 1222 \\
\bottomrule
\end{tabular}
}
\label{table-LA}
\vspace{-0.2cm}
\end{table*}

\section{Details of the Ablation studies}
\label{appendix:ablation-studies}
\subsection{Details of ablation models}
In this section, we explain the details of ScheduleNet variants {\fontfamily{lmtt}\selectfont GN-CR}, {\fontfamily{lmtt}\selectfont GN-PPO}.
Both variants employ the attention GN blocks (layer) to embed $\gG_\tau$. The attention GN block takes a set of node embeddings $\mathbb{V}$ and edge embeddings $\mathbb{E}$, and produces the updated node embeddings $\mathbb{V}'$ and edge embeddings $\mathbb{E}'$ by utilizing three trainable modules (edge function $f_e(\cdot)$, attention function $f_a(\cdot)$ and node function $f_n(\cdot)$) and one aggregation function $\rho(\cdot)$. The computation procedure of the attention GN block is given in Algorithm \ref{alg:gn_block}.

\begin{algorithm}[h]
\caption{Attention GN block}
\label{alg:gn_block}
\SetKwInOut{Input}{input}
\SetKwInOut{Output}{output}
\SetKwInOut{Return}{return}

\Input {
      set of edges features $\mathbb{E}$ \newline
      set of nodes features $\mathbb{V}$ \newline
      edge function  $f_{e}(\cdot)$ \newline 
      attention function $f_{a}(\cdot)$ \newline
      node function  $f_{n}(\cdot)$ \newline
      edge aggregation function $\rho(\cdot)$
      }
$\mathbb{E}',\mathbb{V}' \leftarrow \{\}, \{\}$ \tcp*{Initialize empty sets}
\For{$e_{ij}$ \textbf{in} $\mathbb{E}$}{
    $e_{ij}' \leftarrow f_{e}(e_{ij}, v_{i}, v_{j})$ 
    \tcp*{Update edge features}
    $z_{ij} \leftarrow f_{a}(e_{ij}, v_{i}, v_{j})$ 
    \tcp*{Compute attention logits}
    $\mathbb{E}' \leftarrow \mathbb{E}' \cup \{e'_{ij}\}$    
}
\For{$v_i$ \textbf{in} $\mathbb{V}$}{
    $w_{ji} \leftarrow \textit{softmax}(\{z_{ji}\}_{j \in \mathcal{N}(i)})$
    \tcp*{Normalize attention logits}
    $m_{i} \leftarrow \rho(\{w_{ji} \times e_{ji}'\}_{j \in \mathcal{N}(i)})$
    \tcp*{Aggregate incoming messages}
    $v_i' \leftarrow f_{n}(v_i, m_{i})$
    \tcp*{Update node features}
    $\mathbb{V}' \leftarrow \mathbb{V}' \cup \{v'_{i}\}$
    }
\Return{Updated node features  $\mathbb{V}'$ and edge  $\mathbb{E}'$ features}    
\end{algorithm}

\paragraph{Details of {\fontfamily{lmtt}\selectfont GN-CR}}
\begin{itemize}[leftmargin=*]
\setlength\itemsep{-0.1em}
    \item \textbf{Network architecture.} {\fontfamily{lmtt}\selectfont GN-CR} is composed of two attention GN layers. The first GN layer encodes the raw node/edge features to the hidden embedding. Each GN layer's encoders $f_e$, $f_a$, and $f_n$ has the same parameters as $\text{MLP}_{edge}$, $\text{MLP}_{attn}$, $\text{MLP}_{node}$ respectively, as shown in Table \ref{table-tga-parameters}. We use $\rho(\cdot)$ as the average operator. The second GN layer has the same architecture as the first GN layer but the input and output dimensions of $f_e$, $f_a$, and $f_n$ are 32, 32, and 32 respectively. All hidden actionvations are ReLU, and output activations are indentity function.
    \item \textbf{Action assignments.} Similar to {\fontfamily{lmtt}\selectfont ScheduleNet}, we perform raw feature encoding with the first GN layer and  $H$-rounds of hidden embedding with the second GN layer. We use the same MLP architecture of {\fontfamily{lmtt}\selectfont ScheduleNet} to compute the assignment probabilities from the embeded graph.
    \item \textbf{Training.} Same as {\fontfamily{lmtt}\selectfont ScheduleNet}.
\end{itemize}

\paragraph{Details of {\fontfamily{lmtt}\selectfont GN-PPO}.}
\begin{itemize}[leftmargin=*]
\setlength\itemsep{-0.1em}
    \item \textbf{Network architecture.} Actor (policy) is the same as {\fontfamily{lmtt}\selectfont GN-CR}. Critic (value function) utilize the same GNN architecture to embed the input graph $\gG_\tau$. On the embedded graph, we perform the average readout function to readout the embedded information. All other MLP parameters are the same as {\fontfamily{lmtt}\selectfont GN-CR}.
    \item \textbf{Training.} We use proximal policy gradient (PPO) \cite{schulman2017proximal} to train {\fontfamily{lmtt}\selectfont GN-PPO} with the default PPO hyperparamters of the \texttt{stable baseline} PPO2 implementation. The hyperparameters can be found in \url{https://stable-baselines.readthedocs.io/en/master/modules/ppo2.html}.
\end{itemize}


\begin{figure}[t]
\begin{center}
\includegraphics[width={0.30\linewidth}]{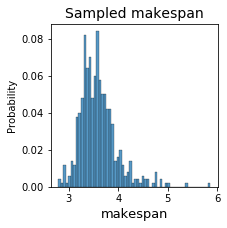}
\end{center}
\caption{\textbf{makespan distribution}}
\label{fig:sampled_gaps}
\end{figure}

\subsection{Extended discussion on Actor-Critic approach.}
As explained in Section \ref{section:mtsp-experiments}, the model trained with PPO  ({\fontfamily{lmtt}\selectfont GN-PPO}) results in lower scheduling performance as compared to the models trained with Clip-REINFORCE ({\fontfamily{lmtt}\selectfont ScheduleNet} and {\fontfamily{lmtt}\selectfont GN-CR}). We hypothesize that this phenomenon is because of the high volatility and multi-modality of the critic's training target (sampled makepsan) as visualized in Figure \ref{fig:sampled_gaps}. This may cause inaccurate state-value predictions of the critic. The value prediction error would deteriorate the policy due to the bellman error propagation in the actor-critic setup as discussed in \cite{fujimoto2018addressing}.

\section{Limitations and future works.}

ScheduleNet is an end-to-end learned heuristic that constructs feasible solutions from ``scratch'' without relying on existing solvers and/or effective local search heuristics. ScheduleNet builds a complete solution sequentially by accounting for the current partial solution and actions of other agents. This poses a challenge for cooperative decision-making with sparse and episodic team-reward. As we use only a single-shared scheduling policy (ScheduleNet) for all agents, it can be effectively transferred to solve extremely large mSPs. We propose ScheduleNet not only to achieve the best performance in mathematically well-defined scheduling problems, but also to address the challenging research question: \textbf{\textit{“Learning to solve various large-scale multi-agent scheduling problems in a sequential, decentralized and cooperative manner”}}. The following are the limitations of the current study and the future research direction to overcome these limitations.

\paragraph{Towards improving the performance of ScheduleNet.} The RL approaches for routing problems can be categorized into: (1) \textit{improvement heuristics} which learns to revise a complete solution iteratively to obtain a better solution; and (2) \textit{construction heuristics} learns to construct a solution by sequentially assigning idle vehicles to unvisited cities until the full routing schedule (sequence) is constructed. The improvement heuristics typically can obtain better performance than the construction heuristics as they find the best solution iteratively through the repetitive solution revising/searching process. However, improvement heuristics require expensive computations than construction heuristics. Moreover, such solution revising processes can be a computational bottleneck when the size of multi-agent scheduling problems becomes larger (i.e., many agents and tasks). In this regard, this study expands the construction heuristics to a multi-agent setting to design a computationally efficient and scalable solver. While focusing on providing general schemes to solve various mSPs, ScheduleNet does not have the most competitive performance than the algorithms that are specially designed to solve specific target problems.

We can also employ the structure of the existing scheduling methods, which eventually attains better solutions after a series of computations. For instance, the learnable Monte-Carlo tree search (MCTS) \cite{guez2018learning} learns the tree traversing heuristics and empirically shows better tree search performance than UBT/UCT based MCTS. We can also borrow the structure of existing scheduling algorithms (e.g., LKH3, Branch $\&$ Bound, $N$-opt) to construct improvement-guaranteed policy-learning schemes.

\paragraph{Towards more realistic mSP solver.} The proposed MDP formulation framework allows modeling more realistic/complex mSPs. Nowadays, the consideration of the random task demand and agent supply are necessitated for real-life mSP applications (e.g., robot taxi service with autonomous vehicles). In this study, we only aim to solve classical mSP problems, which are mathematically well-studied and have baseline heuristics to confirm the potential of the proposed SchdeduleNet framework. We expect that the current ScheduleNet framework can cope with such real-world scenarios by reformulating the state and reward definition appropriately.

\end{document}